\documentclass[12pt]{article}

\usepackage[utf8]{inputenc}
\usepackage[margin=1.25in]{geometry}
\usepackage{times}			

\usepackage[table]{xcolor}	
\usepackage{tipa}			
\usepackage{arydshln}		
\usepackage{tikz}			
\usepackage{tikz-qtree}		
\usepackage{bussproofs}		
\usepackage{stmaryrd}		
\usepackage[linguistics]{forest}	
\forestset{qtree/.style={for tree={parent anchor=south, 
          child anchor=north,align=center,inner sep=0pt}}}
           
\usepackage{multicol}		
\usepackage{multirow}		
\usepackage{graphics}		
\usepackage{adjustbox}		
\usepackage[round]{natbib}			
\usepackage{linguex}		
\usepackage{soul}			
\usepackage{pbox}			
\usepackage{amsmath}		
\usepackage{amssymb}		
\usepackage{pifont}			
\usepackage{outlines}		
\usepackage{stackrel}		
\usepackage{caption}		
\usepackage{hyperref}		
\usepackage{bm}				
\usepackage{hyperref}		
\usepackage{csquotes}		
\usepackage{array}			
\usepackage{booktabs}		
\usepackage{url}			
\usepackage{subfiles}
\usepackage{lscape}
\usepackage{enumitem}
\usepackage[font=small,labelfont=bf]{caption}
\usepackage{tcolorbox}

\author{Alex Warstadt, Samuel R. Bowman}
\date{}
\title{What Artificial Neural Networks Can Tell Us About Human Language Acquisition\footnote{\emph{Publication note}: This paper is published as a chapter of the same title in the volume: \textit{Algebraic Structures in Natural Language}. Shalom Lappin \& Jean-Philippe Bernardy, editors. Taylor \& Francis. \emph{2022}. This document includes minor edits and additions not reflected in the book chapter.\\\emph{Update, February 2024}: This manuscript was revised to correct information about the CHILDES corpora \citep{macwhinney2014childes}. The original article stated that CHILDES contains about 5 million words of American English. However, this figure from \citet{huebner2021using} excludes speech produced by children and all data without child age information, and it also omits the sizable British English corpora. All English corpora including child speech comprise over 20 million words.}}
\begin{document}
\maketitle

\def\todo#1{{\color{red}[TODO: #1]}}
\def\questionfor#1#2{{\color{blue}[question for #1: #2]}}
\def\newterm#1{\textit{#1}}

\section*{Abstract}
Rapid progress in machine learning for natural language processing has the potential to transform debates about how humans learn language. However, the learning environments and biases of current artificial learners and humans diverge in ways that weaken the impact of the evidence obtained from learning simulations. For example, today's most effective neural language models are trained on roughly one thousand times the amount of linguistic data available to a typical child. To increase the relevance of learnability results from computational models, we need to train model learners without significant advantages over humans. 
If an appropriate model successfully acquires some target linguistic knowledge, it can provide a proof of concept that the target is learnable in a hypothesized human learning scenario.
Plausible model learners will enable us to carry out experimental manipulations to make causal inferences about variables in the learning environment, and to rigorously test poverty-of-the-stimulus-style claims arguing for innate linguistic knowledge in humans on the basis of speculations about learnability. Comparable experiments will never be possible with human subjects due to practical and ethical considerations, making model learners an indispensable resource.
So far, attempts to deprive current models of unfair advantages obtain sub-human results for key grammatical behaviors such as acceptability judgments. But before we can justifiably conclude that language learning requires more prior domain-specific knowledge than current models possess, we must first explore non-linguistic inputs in the form of multimodal stimuli and multi-agent interaction as ways to make our learners more efficient at learning from limited linguistic input.

\section{Introduction}

In the 13th century, the Holy Roman Emperor Frederick II conducted a troubling experiment. He arranged for children to be raised from infancy without any human language in order to answer the following question: Which language do children know from birth---Hebrew, Latin, Greek, or their mother's native tongue \citep{coulton1972princes}? Similar experiments were reportedly conducted by the Pharaoh Psamtik and Scotland's King James IV \citep{fromkin1974development}. Despite obvious ethical reasons not to conduct more experiments like this, it is clear that they get at long-standing questions in the acquisition and origins of language that we have few other viable methods of addressing.

\begin{figure}
    \centering
    \includegraphics{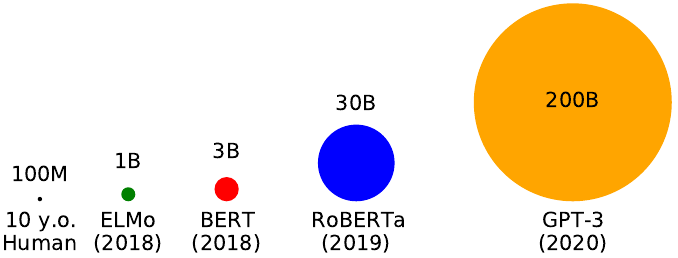}
    \caption{Comparison of human and model linguistic input (\# of word tokens).}
    \label{fig:training_size}
\end{figure}

In the last decade, this possibility has begun to come within reach---without any of the ethical baggage---through the study of artificial neural networks (ANNs). Since the overwhelming success of deep learning methods in natural language processing \citep{manning2015computational,lecun2015deep}, we have gained access to ANNs that learn to compose high-quality multi-paragraph prose, to answer high school-level reading-comprehension questions, and to make human-like grammatical acceptability judgments \citep{devlin2019bert,liu2019roberta,he2020deberta,brown2020language,rae2021scaling,chowdhery2022palm}. These models are all kinds of \emph{language models} (LMs), which learn from unlabeled, naturally occurring text.

In this time, the active research area of \emph{neural network probing} has begun to investigate the grammatical knowledge of LMs \citep{linzen2016assessing,chowdhury2018rnn,gulordava2019colorless,wilcox2018what,warstadt2019linguistic,warstadt2020can,warstadt2020blimp,hu2020systematic,chaves2020what,papadimitriou2021deep,choshen2021grammar}. While these studies collectively find that LMs do not always show human-like language understanding and grammatical intuitions, there has been massive progress in this direction due to both technical innovations and increases in scale over the last few years \citep{manning2020emergent,linzen2021syntactic}. 

Many authors of probing studies suggest that, to the extent that models succeed, this can provide evidence to inform debates about human language acquisition \citep{lau2017grammaticality,warstadt2018neural,chowdhury2018rnn,linzen2019can,pater2019generative,warstadt2020can}. 
However, most studies in this vein focus on convenient but unrealistic learning scenarios, such as large LMs trained on massive corpora scraped from the internet (see Figure \ref{fig:training_size}). As a result, these studies are not well suited to answer questions about \emph{human} language learning, though their methodologies might be a useful stepping stone. At the same time, others have questioned the value of using neural networks to study human language acquisition at all, arguing that their inductive biases are too strong for this to be successful \citep{baroni2021proper}.\footnote{In another piece skeptical about the relevance of LMs to linguistics, Dupre argues that research on LMs is unlikely to provide evidence for discriminating between competing theories of linguistic competence \citep{dupre2021what}. He suggests this is because LMs are trained to simulate human performance, and performance and competence may differ in systematic ways. However, as Dupre himself notes, this argument is ``consistent
with recent work...that has argued that [deep learning] may
provide insight into the mechanisms by which linguistic competence is acquired.'' His argument does not rule out the possibility of inferring the linguistic competence of LMs. While this may currently differ from human linguistic competence too much to provide a testing ground for linguistic theories, we can still use comparisons between LM and human competence to test learnability claims.}

The goal of this chapter is to characterize what we can (and cannot) hope to learn about human language acquisition from studying artificial learners, and how best to maximize the relevance of studies on ANNs to questions of human learning. We agree with many others who contend that artificial neural networks are especially well suited to provide proofs of concept for the possibilities of low-bias learning
\citep{lau2017grammaticality,warstadt2018neural,chowdhury2018rnn,linzen2019can,pater2019generative,warstadt2020can}. We make this claim more precise by arguing that computational modeling results can be made stronger under certain conditions, and that the most easily achievable conditions are the ones for showing that some hypothesized advantage $A$ is not necessary for acquiring some target linguistic knowledge $T$.

\begin{table}[]
    \centering
    \begin{tabular}{ll}
    \toprule
        \textbf{Learner Inductive Bias} & \textbf{Learning Environment} \\
        \midrule
        Architecture & Modality of input\\
        Learning algorithm & Quantity of data\\
        Hyperparameters & Data distribution (domain)\\
        Order of presentation & Memory\\
        \bottomrule
    \end{tabular}
    \caption[Properties of a \newterm{learning scenario}.]{A \newterm{learning scenario} is characterized by the learner and its innate inductive bias on the one hand, and the learning environment on the other. These two factors can be broken down further into several variables, listed non-exhaustively here.}
    \label{tab:learning_scenario_variables}
\end{table}

The way to accomplish this is through an \emph{ablation study} (or deprivation experiment) of a learning scenario which lacks $A$. For our purposes, a \newterm{learning scenario} is determined by two main variables---the innate inductive bias of the learner, and the learning environment (see Table \ref{tab:learning_scenario_variables})---and an \newterm{advantage} is any innate bias or environmental stimulus which is assumed to be helpful for acquiring $T$.\footnote{More precisely, $A$ can be considered an advantage if and only if a learning scenario including $A$ is more likely than a similar learning scenario without $A$ to lead to the successful acquisition of $T$.}
If the model succeeds after ablating $A$, it provides a proof of concept that the target is learnable without $A$. If the model furthermore does not enjoy any substantive advantages over humans, then we can conclude the result is likely to generalize to humans, and considerations from learnability do not justify the claim that humans require $A$.

\begin{tcolorbox}[title=Example: A toy ablation study]

Goal: To test whether learners need to see triply embedded clauses to judge their acceptability under an assumption (which may or may not be correct) that humans learn in a grounded environment without a hierarchical bias. In the terminology of this paper, we test whether $A$ is necessary to learn $T$ under the hypothesis that $S_H$ is the human learning scenario. To do so, we ablate $A$ by removing from the input all examples like (1), and evaluate for knowledge of $T$.

\begin{itemize}
    \item \textbf{Hypothesized Advantage $A$:} Direct exposure to triply embedded clauses (1):
\begin{itemize}
\item[(1)] I think you know that John told the lawyer where \underline{the girl lives}.
\end{itemize}

\item \textbf{Target Knowledge} $T$: The ability to identify agreement errors of triply embedded verbs, e.g.~\emph{live} vs.~\emph{lives} in (1).

\item \textbf{Hypothetical Human Scenario $S_H$:}  \begin{itemize}
    \item[a.] Humans have grounding.
    \item[b.] Humans lack a hierarchical bias.
    \item[c.] ...
\end{itemize}

\end{itemize}

\end{tcolorbox}

Consequently, positive results from model learners are more meaningful than negative results. The example above clarifies why: Supposing we tested a model and found that it acquired the $T$ after ablating $A$. This result is likely to generalize from the model to the hypothesized human scenario as long as the model does not have any additional advantage over what we have hypothesized for humans, i.e. as long as the model also lacks a hierarchical bias. The result is generalizable whether or not the model has grounding, under the assumption that grounding can only help---not hinder---learning.\footnote{For the sake of illustration, we assume in this example that the only advantages besides $A$ where the model can differ from humans are grounding and a hierarchical bias.}
By contrast, to infer that $A$ \emph{is} likely necessary for humans, the model must fail in a scenario that has at least as many advantages as a human, i.e.~it must also have grounding. Otherwise, a lack of grounding would be a potential confound (perhaps exposure to the semantics of number accelerates morphological category learning). While this example is idealized, we believe it is generally far more practical to aim for model learners that lack any unfair advantages over humans, than to try to equip models with the full richness of the hypothesized human learning scenario.

Ablation experiments can provide a rigorous test for claims common in the language acquisition literature that the input to the learner lacks key evidence for acquiring certain forms of linguistic knowledge \citep{chomsky1971problems,legate2002empirical,lidz2003what,berwick2011poverty,rasin2021nature}. They can also refute long-standing hypotheses that certain innate language-specific biases are necessary to explain human language learning \citep{chomsky1965aspects,chomsky1993theory}. The implications for human learning have some limits, though. Showing that some bias is not necessary does not entail it is not present. There are good reasons why something which is learnable may nevertheless be innate: An individual that is born with some crucial knowledge will have an advantage over a similar individual that must acquire that knowledge.

To maximize the probability of obtaining generalizable results from model learners, we need to modify current widely available models to be weaker in some ways and stronger in others. For example, state-of-the-art language models have a huge advantage over humans in terms of the quantity of linguistic input (see Figure \ref{fig:training_size}). However, when there have been attempts to deprive models of this advantage \citep{vanschijndel2019quantity,warstadt2020learning,zhang2021when}, the performance of the models suffers. The most practical course of action to close the data-efficiency gap between neural networks and humans is to exchange some of the massive amounts of text-only input that current LMs learn from with some of the non-linguistic advantages humans enjoy. These include multimodal inputs such as images and video, as well as input from interaction with other agents with adult grammars.

This paper begins with a theoretical discussion of how to apply the evidence from model learners to humans (Section \ref{sec:models_theory}), before turning to more practical considerations. It surveys existing benchmarks and evaluation methods that can be used to test for human-like linguistic performance (Section \ref{sec:tests}). It then reviews ways in which the learning environment and the learner can be (and have been) adapted to improve the relevance of results from model learners to human (Sections \ref{sec:environment} and \ref{sec:learner}). The discussion (Section \ref{sec:discussion}) argues the case for using artificial neural networks as model learners and lays out a path toward building even more relevant models.

\section{Evidence from Model Learners}\label{sec:models_theory}

Models in science are imperfect. The benefit of studying a model is to generalize results from a tractable or observable setting to a more fundamentally interesting setting. Thus, the most useful learnability results from model learners are ones that are likely to imply similar conclusions about humans. From this perspective, not all models or results are created equal. We recommend a strategy in which relatively impoverished model learners are used to obtain proofs of concept for learnability. We justify this recommendation through theoretical considerations about the conditions under which results generalize from models to humans. 

\subsection{Generalizing Learnability Results from Models to Humans}

One goal of the study of language acquisition is to determine the necessary and sufficient conditions for language learning in humans. However, there are some things we cannot easily learn just by observing humans. Observing an advantage in the human learning environment does not tell us whether it is \emph{necessary} for language learning. For example, some children will receive explicit instruction in grammar, including negative evidence, but it is a matter of debate whether such an advantage is necessary or even useful \citep{marcus1993negative,chouinard2003adult}. Furthermore, significant advances in neuroscience are needed before we can directly and confidently observe whether the specific mechanisms in a hypothesized innate language acquisition device are present in the brain. 

Learnability results from a model can inform all of these questions. We can gain evidence about whether negative evidence is necessary to learn some target $T$ by observing whether or not a model learner learns $T$ without it. Similarly, we can test whether some hypothesized inductive bias (e.g., a bias towards hierarchical syntax) is necessary to learn $T$ by evaluating a model without that bias for knowledge of $T$. However, such results only tell us directly whether these advantages are necessary in the model learning scenario, leaving uncertainty about the relevance for humans. 

Unfortunately, there may be significant differences between the learning scenarios of models and humans that make this inductive leap challenging. One solution is to try to reverse-engineer the human learning scenario, as advocated by \citet{dupoux2018cognitive}. As differences between the model and human scenario decrease, there is greater overlap between what is and is not learnable for the model and for humans. This means that as the model becomes more realistic, an arbitrary learnability result from that model is more likely to generalize to humans.

Thus, even imperfect models can provide useful evidence about human language learning. In fact, it is even possible to construct hypothetical scenarios where the success of an imperfect model learner is certain to generalize to a (hypothesized) human scenario. The example below illustrates a difference between model and human learning scenarios that has no effect on the generalization of results:\\

\begin{tcolorbox}[title=Example: A difference that is irrelevant to the generalizability of the  result]
Let the model $M$ be a perfect simulation of a human, with one exception: When $M$ encounters a sentence surrounded by the tags \texttt{<bigram>...</bigram>}, it outputs the probability of the sentence under a bigram model. Any results from studying $M$ generalize with full confidence to humans, as long as sequences with the \texttt{<bigram>...</bigram>} tags are irrelevant to our evaluation of the model's behavior.
\end{tcolorbox}

More interestingly, some differences do not interfere with the intended conclusion because they make the result even stronger. There are two ways this can happen: First, if the model is at a strict disadvantage relative to the hypothesized human scenario, and the model succeeds, then the hypothesized human scenario must also be sufficient for learning the target. Second (and less practical), if the model is at a strict advantage and fails, then the hypothesized human scenario must be insufficient. \\

\begin{tcolorbox}[title=Example: A difference that strengthens the generalizability of the result]
Let $M$ be a model learner that is identical to a hypothetical human. Suppose we are testing the hypothesis that humans need at least 50 examples of a noun with irregular plural marking to learn (up to some threshold for success) whether it is singular or plural. Take a set of irregular plurals which the model learns, and for which the frequency in a typical learning environment is about 50, and remove or replace instances until there are fewer than 50 tokens of each noun. Further suppose the model environment has another unintended disadvantage: There is an implausibly high rate of subject-verb agreement errors for irregular plurals. If the model succeeds in the ablated setting despite an (unrelated) disadvantage, then humans must also succeed.

\end{tcolorbox}

When such differences are present, the generalizability of the result depends on whether the model succeeds or fails.
If the intervention of removing irregular plurals causes the model to fail, the added disadvantage of agreement errors is a confound: Would humans also fail, given the same intervention but without agreement errors?




\subsection{Why Positive Results Are More Relevant}

In practice, the model learners that we will have access to will be imperfect in many ways. But the example above shows that to answer questions about learnability, it is not necessary to aim for perfect models: It is enough to \emph{undershoot} the advantages of human learners. This alone does not explain why positive results are more relevant, because there is a symmetric case: If our models \emph{overshoot} the advantages of human learners, then their failures are likely to generalize to humans.

The reason why positive results are more generalizable is simply that it is easier to build models that undershoot the advantages of humans. Table \ref{tab:advantages} gives a rough overview of the relative advantages of humans and models. While neither has a strict advantage or disadvantage, it is obvious that it will be easier to deprive our models of their current advantages than to equip them with all the advantages of humans. This is especially true of environmental advantages, as we discuss in Section \ref{sec:environment}.

\begin{table}[]
    \centering
    \begin{tabular}{lp{2in}p{2in}}
    \toprule
      &  \bf Human advantages  & \bf Model Advantages \\\midrule
        \bf Environmental & Sensorimotor stimuli\newline Inter-agent interaction \newline Environmental interaction \newline Prosody & Quantity of text\newline Text domain (edited writing) \newline Punctuation \\\midrule
        \bf Innate & Number of parameters \newline Domain-general bias \newline Language-specific bias (?) & Numerical precision \newline Working memory capacity \\
        \bottomrule
    \end{tabular}
    \caption{Relative advantages of humans and typical LMs.}
    \label{tab:advantages}
\end{table}

This suggests that one strategy for obtaining strongly generalizable learnability results is to severely impoverish model learners. The problem with this strategy is that we are unlikely to observe positive results from very weak learners. If we find that our impoverished models fail, the next course of action is to test whether this is really due to the ablation by enriching the model scenario in innocent ways. This could involve adding to the model scenario sensorimotor input, interaction, and other advantages that humans enjoy.

\subsection{Applying Ablations to Debates in Language Acquisition}

The literature on language acquisition has centered around the necessity and sufficiency of innate advantages and environmental advantages. Nativist arguments in favor of a richer set of innate advantages for humans tend to be supported by claims of the insufficiency of certain aspects of the environment, often under the rubric of \emph{poverty of the stimulus arguments} \cite[i.a.]{chomsky1965aspects,chomsky1971problems,baker1978introduction,crain1987structure,legate2002empirical,fodor2002understanding}. Empiricist rebuttals to this position usually try to argue that known advantages in the human learning environment are sufficient to explain learning given domain-general innate bias \cite[i.a.]{landauer1997solution,reali2005uncovering,perfors2011learnability}.

Ablation studies with model learners are best suited to supporting empiricist claims and refuting nativist claims. This is simply because, for reasons discussed above, positive results or proofs of concepts are more practically generalizable than negative results. To get strong evidence for a nativist claim, one would have to show that an ablation leads to failure in a model scenario without significant disadvantages relative to a typical human.

There are many specific phenomena where the input to learners has been argued to be too impoverished to acquire the observed linguistic behavior without some innate language-specific bias. An exhaustive survey of such claims is beyond the scope of this paper, but some examples (mostly discussed in, but not necessarily limited to, English) include subject auxiliary inversion \citep{chomsky1971problems,crain1987structure,legate2002empirical}, other structure-dependent transformations \citep{berwick2011poverty}, plurals in noun-noun compounds \citep{gordon1985levelordering}, auxiliary sequence ordering \citep{kimball1973formal}, anaphoric ``one" \citep{baker1978introduction,lidz2003what}, the denotation of ``every" \citep{rasin2021nature}, epistemic meanings of modals \citep{vandoorenAcceptedfiguring}, binding \citep{reuland2017grammar}, and verb position in Korean \citep{han2016endogenous}.

\section{Tests of Human-Like Linguistic Knowledge}\label{sec:tests}

Ablation experiments require subjecting the model learner to some test of whether it acquires some target human-like linguistic ability. In this section, we discuss in theory what it means to test an artificial learner for human-like linguistic ability in light of the competence/performance distinction, before surveying existing resources that can be used as tests of linguistic performance.

\subsection{Testing for Competence vs. Performance}

In principle, the target linguistic ability we are interested in could be an aspect of human linguistic competence or performance.\footnote{\citet{dupre2021what} discusses the relation between ANNs and competence at length and suggests that ANNs are better viewed as models of human performance rather than competence because they are optimized to reproduce the output of human performance. We broadly agree with this view, and note that it does not contradict our claim that competence for ANNs may still be well-defined and testable.} In practice, however, most tests are behavioral, and therefore target performance. Performance has the advantages of being easily observable and more theory-neutral than competence. By contrast, competence is a theoretical construct even for humans, so a test of competence would always be subject to our degree of belief in the theory. 

We can also study performance to make inferences about competence. We can construe performance very broadly to include many aspects of behavior, ranging from acceptability judgments to order of acquisition and reading time. Although this has its limitations---two systems that have identical behavior in some respects could have very different internal functioning---the more behavioral similarities we observe between two systems, the greater the evidence that they share an underlying mechanism.


Due in part to a massive growth in NLP research focused on LM evaluation and probing, there are now numerous well-motivated, controlled, and challenging tests for different aspects of neural networks' grammatical knowledge. 
These tests fall roughly into two main categories: supervised and unsupervised. Unsupervised tests do not rely on labeled training data or any task-specific training beyond a self-supervised training objective such as language modeling. Thus any linguistic knowledge revealed by these methods can only have been acquired through self-supervised exposure to the learning environment, or to innate abilities of the learner. Supervised tests play a complementary role. While they give models to task-specific instruction not available to humans, supervised tasks can be constructed much like artificial language learning experiments on humans \citep{gomez2000infant} to answer a different set of questions.

\subsection{Unsupervised Tests}

Unsupervised tests of LMs take advantage of the fact that LMs are already trained to estimate the likelihood of an element of sequence $w_i$ from the preceding elements $W_{<i}$, and that these predictions can be used to estimate the likelihood of the entire sequence $W$ as a whole:\footnote{This method has been extended to masked language models like BERT \citep{devlin2019bert} by computing \emph{pseudo likelihood} of a sequence \citep{wang2019bert,salazar2020masked} by sequentially masking one word of the sequence:

\begin{equation*}
    P_{MLM}(W) = \prod_{i=1}^{|W|} P_{LM}(w_i | W_{\backslash i})
\end{equation*}

}

\begin{equation*}
    P_{LM}(W) = \prod_{i=1}^{|W|} P_{LM}(w_i | W_{<i})
\end{equation*}

We survey three tasks that use LM likelihood scores to evaluate grammatical knowledge of LMs without additional supervision: acceptability judgments, reading time prediction, and age-of-acquisition prediction.

\subsubsection{Acceptability Judgments, Minimal Pairs, BLiMP}

Acceptability judgments provide a rich behavioral test for grammatical knowledge. They are the main empirical test for many theories of syntax \citep{schutze1996empirical}, and a vast array of subtle human acceptability judgments have been documented by linguists. Furthermore, for native speakers of a language, knowledge of acceptability is both implicit---i.e. not learned through instruction---and widely shared.

Unsupervised acceptability judgments over minimal pairs (sometimes called \newterm{targeted syntactic evaluation}) have become a widespread evaluation method since their first application to LM probing several years ago \citep{linzen2016assessing,marvin2018targeted}.
This method relies on the assumption that a grammatical sentence $W_\textnormal{good}$ should have greater probability than a minimally different ungrammatical sentence $W_\textnormal{bad}$, in order to say that the LM correctly predicts the contrast in acceptability if and only if

\begin{equation*}
    P_{LM}(W_\textnormal{good}) > P_{LM}(W_\textnormal{bad}).
\end{equation*}

Minimal pairs present several advantages. First, they zoom in on the decision boundary between acceptable and unacceptable sentences. Second, they make it possible to evaluate the ability of models to predict gradient differences in acceptability: Even when single-sentence Boolean acceptability judgments are difficult, such forced-choice preference judgments are highly reproducible \citep{sprouse2013li}. Third, the sentences comprising a minimal pair are generally closely matched in length and unigram probability, which are two determinants of a sequence's likelihood orthogonal to acceptability \citep{lau2017grammaticality}.


\begin{table*}[t]\footnotesize\centering
    
    \begin{tabular}{llp{0.36\linewidth}p{0.38\linewidth}}
    \toprule
    \bf Phenomenon & \bf N & \bf Acceptable Example & \bf Unacceptable Example\\\midrule
    Anaphor agr. & 2 & \emph{Many girls insulted \underline{themselves}.} & \emph{Many girls insulted \underline{herself}.}\\
    Arg. structure & 9 & \emph{Rose wasn't \underline{disturbing} Mark.} & \emph{Rose wasn't \underline{boasting} Mark.}\\
    Binding & 7 & \emph{Carlos said that Lori helped \underline{him}.} & \emph{Carlos said that Lori helped  \underline{himself}.} \\
    Control/raising & 5 & \emph{There was \underline{bound} to be a fish escaping.} & \emph{There was \underline{unable} to be a fish escaping.}\\
    Det.-noun agr. & 8 & \emph{Rachelle had bought that \underline{chair}.} & \emph{Rachelle had bought that \underline{chairs}.}\\
    Ellipsis & 2 & \emph{Anne's doctor cleans one \underline{important} \phantom{x} book and Stacey cleans a few.} & \emph{Anne's doctor cleans one book and \phantom{xxx} \phantom{xx} Stacey cleans a few \underline{important}.}\\
    Filler-gap & 7 & \emph{Brett knew \underline{what} many waiters find.} & \emph{Brett knew \underline{that} many waiters find.}\\
    Irregular forms & 2 & \emph{Aaron \underline{broke} the unicycle.} & \emph{Aaron \underline{broken} the unicycle.}\\
    Island effects & 8 & \emph{Which \underline{bikes} is John fixing?} & \emph{Which is John fixing \underline{bikes}?}\\
    NPI licensing & 7 & \emph{The truck has \underline{clearly} tipped over.} & \emph{The truck has \underline{ever} tipped over.}\\
    Quantifiers & 4 & \emph{No boy knew \underline{fewer than} six guys.} & \emph{No boy knew \underline{at most} six guys.}\\
    Subj.-verb agr. & 6 & \emph{These casseroles \underline{disgust} Kayla.} & \emph{These casseroles \underline{disgusts} Kayla.}\\
    \bottomrule
    \end{tabular}
    \caption[Examples from BLiMP.]{Minimal pairs from each of the twelve categories covered by BLiMP. Differences between sentences are underlined. \textit{N} is the number of minimal pair types within each broad category. Table from \citet{warstadt2020blimp}.}\label{tab:phenomena table 1}
    \end{table*}

BLiMP \citep[The Benchmark of Linguistic Minimal Pairs;][]{warstadt2020blimp} is the largest-scale resource for language model scoring. It tests 67 minimal pair types in English, each consisting of 1k pairs, organized into 12 broad categories. These categories cover morphology (e.g.~subject-verb agreement and determiner-noun agreement), syntax (e.g.~argument structure, island effects, and binding), and semantics phenomena (e.g.~quantification and negative polarity items). Table \ref{tab:phenomena table 1} shows examples from BLiMP of one minimal pair type for each of these categories. Closely related is SyntaxGym \citep{gauthier2020syntaxgym,hu2020systematic}, which adopts a version of the LM scoring paradigm in which the model's predictions must conform to more than one hypothesized inequality over a set of sentences, rather than just a minimal pair. 

\subsubsection{Other Behavioral Predictions: Reading Time, Age-of-Acquisition}

Language model scores can be used to predict other aspects of human linguistic performance. Reading times is a prime example. For example, \citet{wilcox2021targeted} test LMs' predictions against humans' online processing difficulty using SyntaxGym. Under the theoretically motivated assumption that there should be a log-linear relationship between a word's online processing time in humans and an LM's predicted probability for the word in context \citep{hale2001probabilistic,levy2008expectationbased}, it is possible to test the conditions under which human-like processing can be acquired. Related is the Natural Stories corpus \citep{futrell2021natural}, which is a benchmark that provides human reading times for a diverse set of sentence types in naturalistic contexts.

Predicting age-of-acquisition is another possible point of comparison between humans and models. Through databases like Wordbank \citep{frank2017wordbank}, we have large-scale multilingual parent-reported data on vocabulary development in a large number of individuals. This data can be used to construct item-level learning trajectories for humans to which we can compare the learning trajectories of LMs \citep{chang2022word}.

\subsection{Supervised Tests}

Supervised classification tasks such as part-of-speech tagging, dependency arc labeling, and coreference resolution have been used as \emph{probing tasks} for model evaluation in NLP \citep{ettinger2016probing,shi2016syntax,adi2017fine,tenney2019you,hewitt2019structural,belinkov2019analysis}. Recently, these methods have been widely criticized because the use of supervision makes it difficult to distinguish knowledge acquired through training on the more cognitively plausible LM task from knowledge acquired through subsequent task-specific fine-tuning \citep{hewitt2019designing,pimentel2020informationtheoretic,voita2020informationtheoretic}. In this section, we survey a family of evaluation tasks that use constrained supervision to probe how neural networks generalize. In this approach, what is under investigation is not knowledge of a particular phenomenon in the training data, but whether models extend knowledge to unseen cases in ways that we expect humans to.

This approach can tell us the extent to which models form generalizations governed by consistent, high-level rules. For example, COGS \cite[Compositional Generalization Challenge based on Semantic Interpretation;][]{kim2020cogs} is a semantic parsing dataset in which certain semantic configurations in the test data are systematically held out from the training data. If a model is able to learn that semantics, syntax, and surface form are related by a set of general compositional and phrase-structure rules, then it should correctly parse a noun in any syntactic position, even if it has only seen that noun in object position during training. 

\begin{figure}
    \centering
    \includegraphics[width=\textwidth]{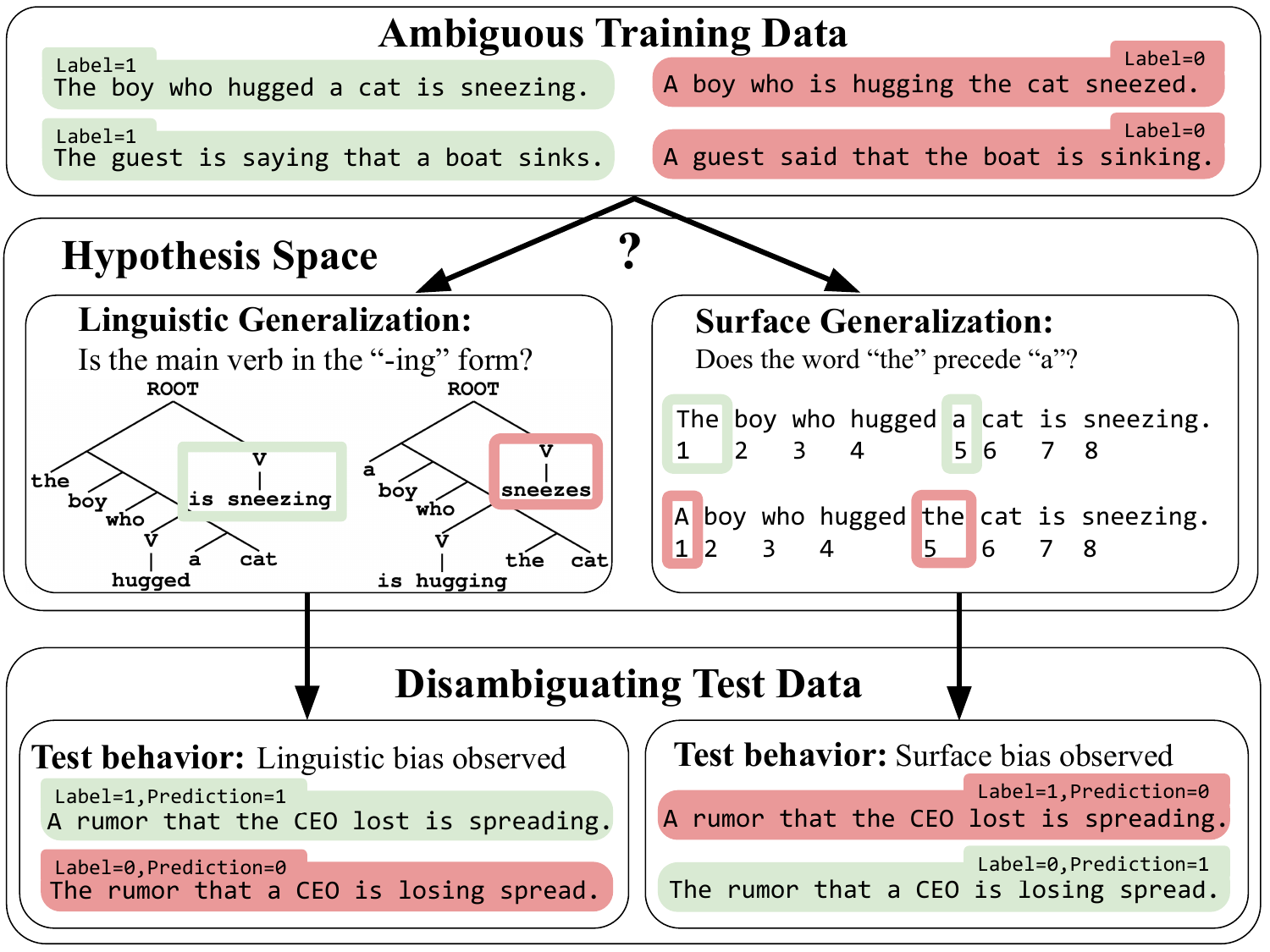}
    \caption[Example of the Poverty of the Stimulus experimental design.]{Example of an experiment following the Poverty of the Stimulus design from the MSGS dataset \citep{warstadt2020learning}. A model is trained on ambiguous data whose labels are consistent with either a linguistic or a surface generalization, and tested on disambiguating data whose labels support only the linguistic generalization. Light green and darker red shading represents data or features associated with the positive and negative labels/predictions, respectively.}
    \label{fig:msgs}
\end{figure}

Constrained supervision is also useful for probing the inductive biases of neural networks. The Poverty of the Stimulus experimental design \citep{wilson2006learning} provides a paradigm for doing so. Figure \ref{fig:msgs} gives an example from \citet{warstadt2020learning} of an experiment following this design. A learner is trained to perform a task given data that is ambiguous between (at least) two hypotheses, and tested on data where the hypotheses make divergent predictions. For example, numerous studies have used this design to test whether ANNs prefer a generalization based on syntactic structure or one based on linear order for subject auxiliary inversion \citep{frank2007transformational,mccoy2018revisiting,mccoy2020does,warstadt2020can}. 

One large-scale dataset making use of this design is MSGS \citep{warstadt2020learning}, which tests whether a learner has a bias towards linguistic or surface generalizations. MSGS consists of 20 ambiguous tasks, each pairing one of four linguistic generalizations (e.g. \emph{labels indicate whether the main verb of the sentence is in the progressive form}) with one of five surface generalizations (e.g. \emph{labels indicate whether the sentence is longer than 10 words}). In concurrent work, \citet{lovering2021predicting} introduced a similar dataset in which linguistic generalizations do not apply linguistic features in arbitrary ways, but correspond to acceptability judgments.

\subsubsection{What Do Out-of-Domain Tests Tell Us About Learnability?}

Although the input to humans is not annotated with linguistic features, training and testing models on a supervised task with such labeled data can still provide useful evidence about human learning. The key is to use LMs to provide task-specific models with linguistic features acquired from a general pretraining setting. This could be an application of the popular pretrain-and-fine-tune paradigm in NLP \citep{dai2015semisupervised,howard2018universal,radford2018improving,devlin2019bert}, or the prompt-based few-shot learning paradigm \citep{brown2020language}.

Following this setup, the experiment can tell us whether an inductive bias, such as a hierarchical bias or a compositionality bias, can be \emph{acquired} through exposure to the unstructured learning environment \citep{warstadt2020learning,warstadt2020can}. An acquired inductive bias, though not present innately in the learner, can still influence how the learner forms generalizations about sub-problems encountered during the learning process. 

\section{The Learning Environment}\label{sec:environment}

The learning environments of large language models diverge from the human learning environment in ways that give both advantages and disadvantages. On the one hand, today's widely studied LMs are exposed to hundreds or thousands of times more words than a child, and much of that text is written or edited. On the other hand, children learn in a grounded environment through interaction with other agents. These are just the most obvious in a long list of differences in the learning environment that are likely to affect language learning. 

The presence of these differences substantially weakens our ability to generalize results from model learners to humans. To achieve strong positive evidence about the conditions for learnability that generalize to humans, it is necessary to create a learning environment for the artificial learner that represents a lower bound on the richness of the input to human learners. The learner's environment should not exceed the quantity or quality of data available to humans. Of course, if a model successfully learns some target knowledge in an environment that is \emph{far poorer} than a human's---for instance, one containing only a few thousand words---this result would be highly likely to generalize to humans. However, initial experiments attempting this by limiting text data quantity to LMs have found that they fail to acquire key linguistic abilities \citep{vanschijndel2019quantity,zhang2021when}. Fortunately, there is ample room to enrich the learning environment of LMs using multimodal inputs and interactive objectives, all without exceeding the richness of the input to humans.


\subsection{Data Quantity}

\begin{figure}
    \centering
    \includegraphics[width=\textwidth]{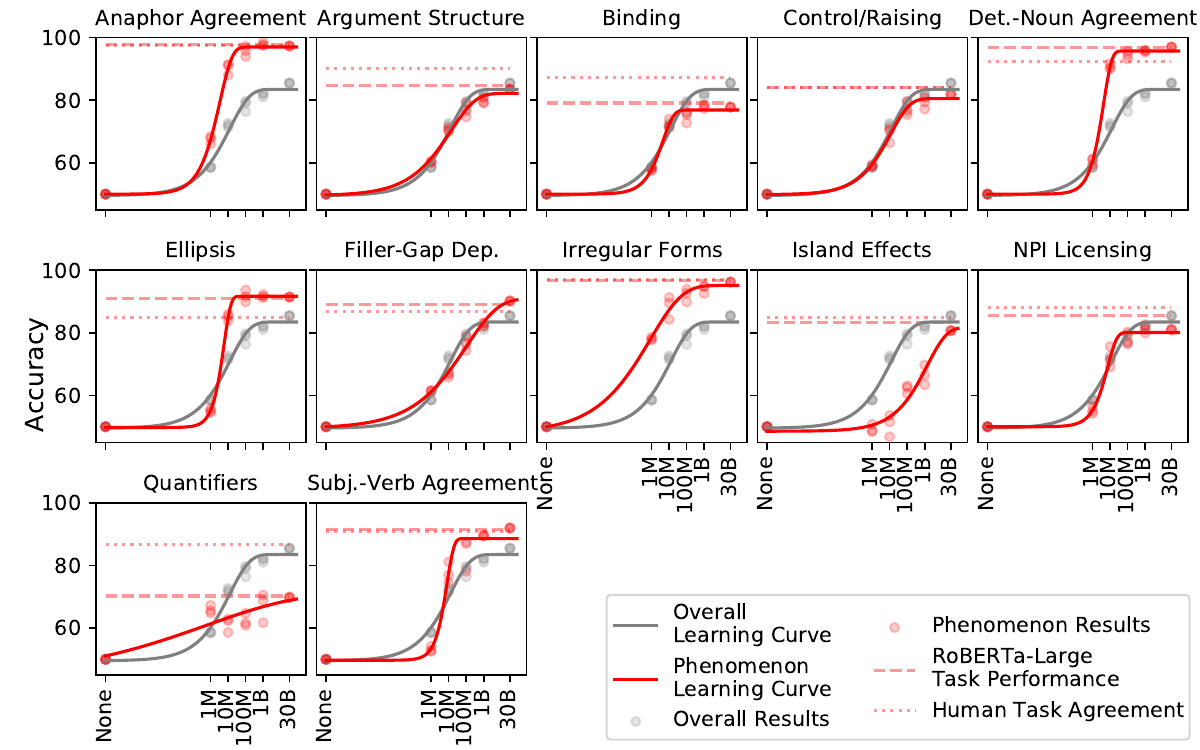}
    \caption[BLiMP performance w.r.t.~training data quantity.]{Learning curves adapted from \citet{zhang2021when}, showing LM improvement in BLiMP performance as a function of the number of words of training data available to the model.}
    \label{fig:blimp_curve_ch1}
\end{figure}

\begin{figure}
    \centering
    \includegraphics[width=\textwidth]{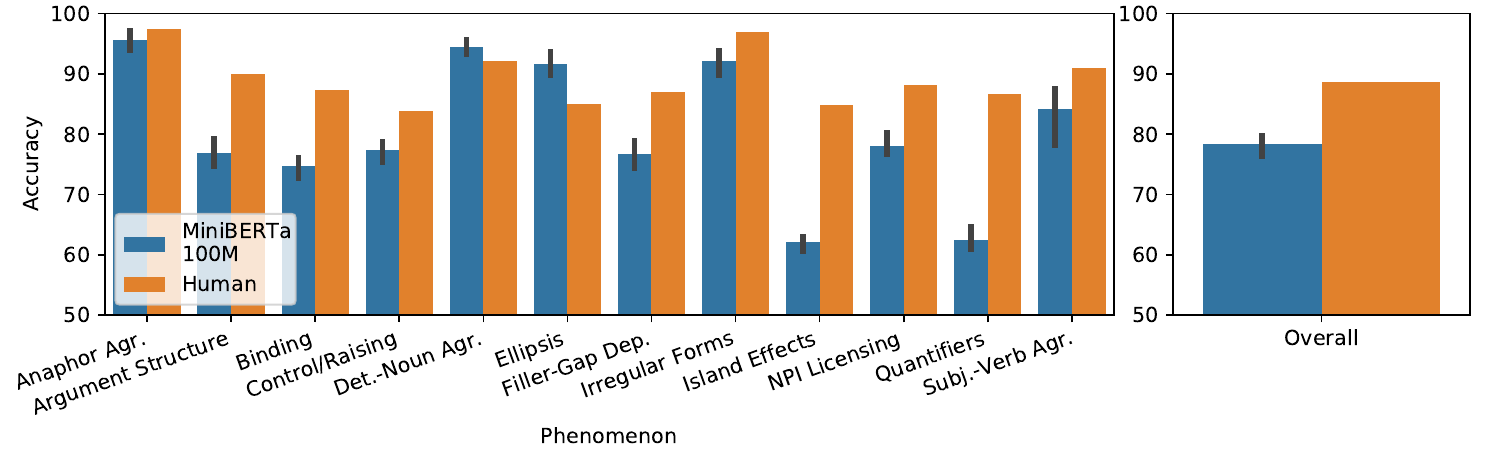}
    \caption[Adult human vs.~human-scale LM performance on BLiMP.]{Comparison of BLiMP performance between adult humans and human-scale LMs. Model results are averages over three 100M word miniBERTas reported in \citet{zhang2021when}.}
    \label{fig:human_model}
\end{figure}

Most popular ANNs for NLP have been trained on far more words than a human learner. While this was not the case only a few years ago, this trend has only been increasing. Thus, researchers interested in questions about human language acquisition have already begun to intentionally shift their focus to evaluating models trained on more human-scale datasets \citep{vanschijndel2019quantity,hu2020systematic,pannitto2020recurrent,warstadt2020blimp,warstadt2020learning,perez-mayos2021how,zhang2021when}.

However, it is not trivial to determine how many words a typical human learner is exposed to. The best-known figures come from Hart \& Risley's study on American English-speaking children's linguistic exposure in the home \citep{hart1992american}. They find that children are exposed to anywhere from 11M words per year to as little as 3M words. These figures include all speech in the home environment, not just child-directed speech. More recent work by \citet{gilkerson2017mapping} places this estimate between approximately 2M and 7M words per year (extrapolated from mean daily words $\pm$ 1 standard deviation). Choosing the beginning of puberty as a rough cutoff point for language acquisition, and assuming that these rates are consistent across childhood, a child will acquire language with anywhere from tens of millions of words to as much as a hundred million words.

By comparison, popular neural language models are trained on corpora consisting of far more data (see Figure \ref{fig:training_size}): ELMo \citep{peters2018} is trained on 1 billion words, BERT \citep{devlin2019bert} is trained on about 3.3 billion words, RoBERTa \citep{liu2019roberta} is trained on about 30 billion, and GPT-3 \citep{brown2020language} is trained on about 200 billion. Thus, the most impoverished of these models has linguistic experience equivalent to about 300 human years, and the most enriched is at 20,000 human years. 

We can already begin to draw some conclusions about how linguistic performance of LMs scales with the quantity of available data. \citet{zhang2021when} uses BLiMP to evaluate models trained on datasets ranging from 1M words to 1B words. Figure \ref{fig:blimp_curve_ch1} summarizes their results, showing the growth in sensitivity to acceptability contrasts as a function of the amount of training data available to an LM. 
They find that language models do learn many human-like generalizations given abundant data when tested using unsupervised LM scoring. RoBERTa$_\textsc{base}$ which is trained on about 30B words \citep{liu2019roberta} achieves near-human performance (which we define as accuracy within 2\% points of humans or better) on 6 out 12 BLiMP categories. Among these categories are phenomena involving long-distance syntactic dependencies such as filler-gap dependencies and island effects, which have been previously found to be challenging for LMs \citep{warstadt2020blimp}.

On the other hand, language models generally fail to reach human-level accuracy when restricted to human-scale data quantities. According to the same study, RoBERTa models trained at human scale on 100M words only achieve near-human performance in at most 2 BLiMP categories (Figure \ref{fig:human_model}). Models trained on 10M words are unsurprisingly even worse, reaching near-human performance in only a single BLiMP category.


\subsection{Data Source}

Another point of divergence between human and model learning environments is the source of language data. One of the main distributional differences is that all the linguistic input to a pre-literate child is spoken or signed. Ideally, the model learner's environment should consist of unstructured audio or video of real-life communication. While there have been some first steps toward training LMs on such data \citep{nguyen2020zero,lakhotia2021generative,lavechin2022early}, these models are not yet advanced enough to learn complex syntax.\footnote{For example, current state-of-the-art performance of audio-trained LMs on a modified audio version of BLiMP is only 58\% accuracy---just 8\% points above chance---compared to over 79\% accuracy for RoBERTa models trained on 100M words.} 

As long as text-based training remains the main viable option for training effective LMs, the most ecologically valid text domain is transcribed speech. 
One source of such data is CHILDES, a database of transcribed parent-child discourse \citep{macwhinney2014childes}. 
Indeed, such infant-directed speech is a major source of input to many child learners, and some go so far as to train model learners exclusively on child-directed speech \citep{reali2005uncovering,perfors2011learnability}.
This is probably overkill: Child-directed speech makes up only a part of the linguistic input to child learners, and in some communities it is vanishingly rare \citep{cristia2019child}. 
Additionally, the volume of data in CHILDES is on the low end of what can be used to train developmentally plausible models if used as the exclusive data source.
All American and British English sources, including both child-directed speech and child utterances, comprise about 20 million words,\footnote{We thank Brian MacWhinney (p.c.) for providing this figure.} equivalent to the amount of linguistic input to an average 4- or 5-year-old. 
Therefore, depending on the intended application for a given model learner, the most developmentally plausible training dataset may include CHILDES in conjunction with other data sources.

Another large-scale source of transcribed speech is COCA \citep{davies2009385}, which includes 83M words of transcribed speech from unscripted radio and TV programs. One step down in terms of ecological validity is OpenSubtitles \citep{lison2016opensubtitles2016}, which contains over 2B words of English subtitles from scripted and unscripted television and radio, as well as over 100M words of subtitles in numerous other languages. While these datasets are ultimately not what is needed to obtain the most generalizable proofs of concept, they can give more compelling evidence than training datasets currently used to train popular language models such as Wikipedia, news, and web data.


\subsection{Prosody}

There is substantial linguistic information in speech not present in text, especially prosody. Prosodic bootstrapping is thought to play a major role in syntactic acquisition \citep{gleitman1982language,soderstrom2003prosodic}, so LMs are at a distinct disadvantage in this respect. On the other hand, text data has punctuation and white space, and is tokenized prior to input into an LM, which provides an \emph{advantage} to models when it comes to detecting word, phrase, and sentence boundaries. Again, if practical limitations are not an issue, it is best to study models trained mostly on audio. But since this is not totally practical at the moment, there is still a lot to learn from LMs trained on text. Text exceeds the richness of speech in fairly limited ways, meaning that results from text-trained LMs still give suggestive evidence about humans. 

\subsection{Non-Linguistic Input}

Despite some advantages in the linguistic environments of typically studied ANNs, they have severe non-linguistic disadvantages compared to humans. Whereas most ANNs studied in this literature learn in a text-only environment with a simple LM training objective, humans learn in a multifaceted environment with many forms of sensory input, other agents to interact with, and complex risks and rewards. The effects of these differences in non-linguistic input on grammar learning are likely to be more indirect than changes in linguistic input. Still, they may turn out to be substantial, especially when it comes to the quantity of linguistic input the learner requires. 

\subsubsection{Multimodal Input}

Theories of language acquisition have long hypothesized a substantial role for sensorimotor input. The presence of a conceptual scaffolding acquired through sensorimotor input has been suggested to accelerate or improve grammar learning \citep{howell2005model}. This is one explanation for the well-known noun bias in early word learning: Concepts for objects may be learned earlier than concepts for \emph{relations} or \emph{properties} of objects \citep{gentner1982why}. 

Any conceptual scaffolding that typical language models possess must be acquired from text alone. Consequently, conceptual knowledge is not available to facilitate the early stages of language learning in LMs.
Indeed, these models can eventually acquire some semblance of world knowledge. Pretrained models can function as knowledge bases for retrieval of encyclopedic knowledge, accurately completing factual statements like \emph{iPod Touch is produced by \underline{\hspace{2em}}} \citep{petroni2019language}, and they achieve strong performance on challenge sets focusing on physical and social common sense \citep{zellers-etal-2018-swag,huang-etal-2019-cosmos,sakaguchi2020winogrande}. But the same models that pass these benchmarks still show signs of inconsistent or contradictory knowledge \citep{elazar2021measuring}. Furthermore, the quantity of training data needed to achieve strong performance on even these limited benchmarks is on the order of billions of words \citep{zhang2021when}. Thus, whatever limited world knowledge language models can acquire is not likely to be useful for language acquisition from human-scale data.

An ideal model learner would experience sensorimotor input that is indistinguishable from a typical child's. If we focus on just the audiovisual domain, the closest usable learning environment to this ideal is the SAYCam dataset \citep{sullivan2021saycam}, which consists of first-person perspective audio and video from head-mounted cameras worn by children under 3 years. While this data has been used for training computer vision models \citep{orhanselfsupervised}, with only an estimated 1-2M words in its audio recordings, it contains too little linguistic data at present to be used to model anything beyond the first few months of language learning.

At the practical end of the spectrum, there is a growing inventory of ANNs trained jointly on vision and language data \citep{lazaridou2015combining,lu2019vilbert,tan2019lxmert,chen2020uniter,Su2020VL-BERT:,radford2021learning,kamath2021mdetr,singh2021flava}. Transformer-based multimodal models accept inputs as text-image pairs, which are passed into a shared multimodal encoder (possibly after passing through separate unimodal encoders). Most are pretrained using self-supervised objectives similar to language modeling. We summarize three representative objectives:
\begin{enumerate}
    \item Masked multimodal modeling: The objective is to reconstruct masked text or image regions from an image-text pair \citep{tan2019lxmert}. Approaches vary as to whether masking occurs only in the text, only in the image, or both.
    \item Image-text matching: The objective is to classify an image-text pair as matched (e.g., an image and its caption) or unmatched (i.e., randomly aligned).
    \item Contrastive: Given $N$ matched image-text pairs, the objective is to jointly maximize the representational similarity of all $N$ matched pairs and minimize the similarity of all $N(N-1)$ mismatched pairs in a shared embedding space \citep{radford2021learning}.
\end{enumerate}

Despite rapid progress in this area, hardly any results so far show that enriching the visual environment of neural networks leads to better language learning. The language encoder of multimodal models is often initialized with weights from a pretrained language-only model, but these multimodal models consistently perform \emph{worse} on linguistic evaluations than the original language-only model \citep{iki2021effecta}. Similarly, models trained end-to-end on a multimodal corpus failed to significantly outperform models trained on the text-only portion of the corpus \citep{yun2021does}. 

These results may reflect limitations of current multimodal models, rather than fundamental limitations of the utility of multimodal input. For example, the linguistic input of typical multimodal models is even farther from that of human learners than language-only models. Most are trained entirely on image caption datasets such as MS COCO or Visual Genome \citep{chen2015microsoft,krishna2017visual}, which lack extended discourses and dialogues and contain a non-representative sample of sentence types. Furthermore, visual input to humans is continuous and moving, and thus richer than still images. Video and language models do not achieve a more realistic training environment. For example, VideoBERT is trained on YouTube cooking videos and text from automatic speech recognition \citep{sun2019videobert}.

\subsubsection{Interactive Learning}

Another ingredient missing from the input to most available model learners is interaction with an environment containing other conversational agents. While the objective of LMs is to reproduce the distribution of words and phrases in the language as faithfully as possible, human learners have a much more complex and varied objective function. We use language to share information, to make queries, and to issue and comprehend directives \citep{austin1962how,searle1969speech}. The incentive for acquiring grammar in humans is that it leads to communicative success in these kinds of interactions, helping us achieve our non-linguistic goals. 

The artificial learning paradigm that comes closest to reproducing this aspect of the human learning environment is multiagent reinforcement learning \citep{lazaridou2017multiagent}. In this framework, multiple artificial agents must develop a mode of communication to meet a cooperative goal, such as solving a reference game. However, the goal of this research, rather than to build more cognitively plausible or efficient model learners, is generally to study language emergence, as the emergent modes of communication differ greatly from human language \citep{kottur-etal-2017-natural,bouchacourt2018how,lazaridou2020emergent}. A good deal of work has gone into engineering agents, environments, and populations \citep{mankewitz2021multi,chaabouni2022emergent} that derive basic properties of human language, such as compositionality \citep{lazaridou2018emergence,ren2019compositional,resnick2020capacity}, communicative efficiency \citep{chaabouni2019anti,rita2020lazimpa}, and learnability \citep{li2019ease,chaabouni2022emergent}.

Still, efforts to incorporate multi-agent interaction and language modeling have been limited. \citet{lazaridou2020multiagent} explore several ways of doing so, by initializing agents with LMs pretrained on natural language, or by training agents on the language modeling and interactive objectives in a multi-task setting. However, these approaches are susceptible to language drift, resulting in communication protocols that are no longer understandable by humans. Thus, more progress needs to be made in combining these objectives before we should expect to see interactive learning closing the data-efficiency gap between model and human learners.

\section{The Learner}\label{sec:learner}

In this section we consider the final condition on strong evidence of learnability: an appropriate model learner. At the theoretical level, the considerations are the same as with the learning environment. Namely, the less the model learner has an advantage over humans (independent of the experimental manipulation), the greater the chance a positive result from an ablation will generalize to humans. In this case, the relevant advantages are properties that are built into the learner, via its architecture or learning algorithm.

However, the path toward building an ideal model learner is far from clear. This contrasts sharply with the situation with learning environments. Determining what constitutes an innate advantage raises theoretical questions about the nature of inductive bias. The task of probing the inductive bias of models is itself a challenging empirical problem. Our ability to control the inductive bias of our model learners is extremely limited, and to truly prove that the inductive bias of a model learner is no more advantageous than that of a human, we need a certain understanding of the innate advantages of humans. Thus, we must rely on our theoretical understanding of human and model learners and empirical results about their biases to make informed---if subjective---judgments about their relative advantages.

\subsection{Formalizing Innate Advantage}

Before tackling these issues, we need a way to quantify and compare innate advantages across humans and models. We can reinterpret the notion of an innate advantage in terms of inductive bias. The inductive bias of a learner determines which generalization it arrives at from a finite set of examples; in other words, how it makes an inductive leap. Roughly, a learner has an innate advantage if its inductive bias favors the ``right'' kind of generalization. 

We can make this intuition more precise. First suppose we are subjecting the learner to a particular evaluation task, in which the objective is to learn a binary classification function for a target concept $C$ over some instance space $X$. For instance, $X$ might be the set of all strings, and $C$ the set containing all and only the acceptable sentences of $X$. Or $X$ might be the set of all ordered pairs of sentences $(s_1, s_2)$, and $C$ is the set of all pairs where $s_1$ is more acceptable than $s_2$. The hypothesis space $\mathbf{H}$ is the set of all binary classification functions $h$ defined over $X$, i.e. the set containing the characteristic function for each element of $\mathcal{P}(X)$.

Recall our goal: To quantify how strongly a learner favors the right kind of generalization such that we can make comparisons between learners. Intuitively, this corresponds to how much prior probability the learner assigns to the classification function $h^*$ that characterizes the target concept $C$:

\begin{align*}
    A(L, h^*) &:= P_L(H=h^*)
\end{align*}

The learner's prior $P_L(H)$ can be better understood by specifying a prior over the sorts of learning environments $e$ that we expect the learner to be exposed to:\footnote{It is not totally clear what kind of distribution to use in realistic experiments, and furthermore, the choice of the distribution can have a large impact on this quantity. However, a useful notion of a prior over hypotheses should be based on environments that support many types of hypotheses related to but not necessarily identical to the target hypotheses. Furthermore, adversarial environments (like the ones considered by \citet{gold1967language}) should have low or zero probability.}

\begin{align*}
    P_L(H) &= \sum_{e \in E} P_L(H, e)\\
          &= \sum_{e \in E} P_L(H|e) \cdot P(e)
\end{align*}

If we assume that $L$ learns by a deterministic algorithm, then $P_L(H|e) = 1$ for a single hypothesis $h \in H$, and $0$ for all others, i.e.~$L$ is a function from a learning environment $e$ to a hypothesis, which we denote $L(e)$. Thus, we can express the learner's innate advantage as follows:

\begin{align*}
    A(L, h^*) &:= P_L(H=h^*)\\
            &= \sum_{e \in E} P_L(H=h^*|e) \cdot P(e) \\
            &= \sum_{\substack{e \in E\\s.t. L(e)=h^*}} P(e) \\
\end{align*}

In other words, the learner's innate advantage is the total probability that it will converge on the target hypotheses, assuming a particular distribution over learning environments. Relativizing advantage to the distribution over environments is motivated by the intuition that some learning environments are more typical than others. A learner does not have an advantage if it assigns higher weight to the correct generalization in certain highly contrived environments, but rarely in typical ones. 
This means that to assess whether one learner has an advantage over another, we must compare advantages using the same prior over learning environments.

One more refinement to the naive notion of innate advantage is justified: There is generally not one classification function $h^*$ that we would accept as human-like, but a whole set of functions $\mathbf{H^*}$. This set could be defined as in the probably approximately correct learning framework \citep{valiant1984theory,haussler1990probably} as the set of generalizations which have generalization error (compared to $h^*$) less than some \emph{error tolerance} $\epsilon$. Or, recognizing the existence of individual variation in adult human grammars, it can be the set of generalizations consistent with being a typical native speaker of the language.

These ingredients allow us to quantify the innate advantage $A$ of learner $L$ relative to the class of target generalizations $\mathbf{H^*}$ as follows:

\begin{equation*}
    A(L, \mathbf{H^*}) := \sum_{\substack{e \in E\\L(e)\in \mathbf{H}^*}} P(E=e).
\end{equation*}

This quantity is quite simply the total probability that $L$ will converge to an acceptable generalization in a typical learning environment (as defined by the prior over environments).

\subsection{A Lower Bound on Human Inductive Bias}

The benefit of a formal definition of innate advantage is that it can provide a rough guideline for determining an appropriate model learner. It does not provide a usable criterion for guaranteeing that the model is suitable, for the simple reason that it is impractical to measure in models and in humans. It also does little to clear the path towards building better model learners. Our ability to control the inductive bias of model learners is limited. We are limited by the available set of learners, and developing effective new artificial learners is a large and mature field of research in its own right. 
Thus, while the theoretical considerations around idea model learners and much like those we considered for model environments, there is far less we can do in practice to guarantee or achieve a tight lower bound on human innate advantage than on the human learning environment.

Comparing the inductive bias of neural networks and humans may require substantial empirical work that is more intensive than conducting an experimental manipulation like an ablation. For a model learner, the naive approach to estimating $A(L, H^*)$ is a Monte Carlo approximation \citep{wilson2020bayesian}, to train the learner repeatedly in sampled learning environments and test on the target evaluation. But this can hardly be a \emph{precondition} for doing an experiment on a model learner, since it entails doing the entire experiment repeatedly. The situation is even worse for measuring human inductive bias. In order for the argument from an ablation to go through, we must convince ourselves that humans have at least as strong an advantage in the ablated environment. To determine this, we would need to estimate a human's prior distribution over generalizations \emph{in the ablated environment}. But if we already had this information, this would eliminate much of the need to study model learners in the first place. The only convincing solution to this problem would be to develop techniques to compare inductive bias in models and humans without relying on observations of how each generalizes in typical environments.

\subsubsection{Misconception 1: A Good Model Learner Must be Unbiased}

One possible misconception is a model learner must be an unbiased \emph{tabula rasa} in order to prove some innate bias unnecessary for language acquisition. First, this would be an impossible standard to meet, since all learners have some inductive bias. An inductive bias is just a prior over the hypothesis space, and thus a necessary property of any learner \citep{mitchell1980need}. Second, we know of no claims that humans are totally unbiased learners.
Many do argue that \emph{language-specific} biases are not necessary to explain language acquisition \citep{kirby1999function,reali2005uncovering,clark2011linguistic,christiansen2016creating}. They suggest that we may instead have innate \emph{domain-general} biases which aid us in language acquisition. For an existence proof of this claim, the model learner only needs to lack language-specific bias, and can possess domain-general bias as long it is no stronger than a human's.

What does it mean for a bias to be language-specific? It is not clear that this is even a precise notion. An example of the subtlety of this issue is hierarchical bias. Chomsky famously argues that humans have a bias towards forming a generalization based on syntactic structures about grammatical operations like subject-auxiliary inversion, when linear generalizations would adequately describe most of the data \citep{chomsky1965aspects}. However, it is possible to question how language-specific even this bias is, since non-linguistic aspects of human cognition also make use of hierarchical structures, such as music \citep{lerdahl1983generative} and categorization. More recently, Chomsky has claimed that the primary innate endowment that enables language learning is unbounded Merge, or the ability to form recursive concepts \citep{chomsky2007biolinguistic}. Merge in this view emerged prior to language as we know it: It would have evolved mainly to facilitate abstract thought, with language later co-opting this operation. While Chomsky suggests that Merge in this incarnation is implicated in the language of thought, whether it can be claimed to be truly language-specific seems to be a matter more of terminology than of actual substantive debate. 
Ultimately, it may be misguided to think of inductive bias as language-specific or not. Instead, learners that place more prior probability on generalizations that make use of the grammatical structures we believe to be present in natural language have a stronger linguistic bias.

\subsubsection{Misconception 2: More Expressive Models Have an Advantage}

Another possible misconception is that a learner with greater expressive capacity has an advantage compared to a less expressive one. Of course, this is often the case: There are many examples of more expressive models having an advantage over less expressive ones. For example, a unigram LM is less expressive than a bigram LM with backoff \citep{katz1987estimation}, and it clearly has a disadvantage when it comes to modeling some domains of grammar, such as local subject-verb agreement in strings like \emph{The cats purr}/\emph{*The cat purr}. But this is due to the fact that the bigram LM, but not the unigram LM, reasons over a hypothesis space that overlaps with the set of target generalization $\mathbf{H}^*$, while the unigram model does not. In other words, $A(\textit{unigram}, \mathbf{H}^*) = 0$, and $A(\textit{bigram}, \mathbf{H}^*) \geq 0$. We get the impression that less expressive models may be generally at a disadvantage simply because a smaller hypothesis space often has less overlap with $\mathbf{H}^*$.

But in fact, a learner can become sometimes more advantaged by becoming less expressive. This happens when the learner's hypothesis space shrinks in a highly specific way to exclude incorrect hypotheses, and some of the freed probability mass is placed on $\mathbf{H}^*$. This is precisely the nature of the innate advantages hypothesized in nativist theories of language acquisition.

A similar kind of innate advantage is often built into Bayesian models of language acquisition \citep{perfors2011learnability,abend2017bootstrapping,yang2022one}. To focus on one example, Perfors et al.~study a Bayesian grammar induction system that reasons over a hypothesis space consisting only of a few types of formal systems, including a (limited) set of context-free grammars and finite-state grammars. Compared to a typical LSTM or Transformer language model, this learner has a highly restrictive and peaked prior. While this would be a disadvantage for many kinds of tasks, it is an advantage when it comes to learning specific rules of English syntax, because the learner's inductive bias places relatively great weight on English-like grammars.

\subsection{Achieving a Lower Bound on Human Inductive Bias in Practice}

Practically, our ability to choose appropriate model learners is constrained by the available models. In recent years, our understanding of these models' inductive biases has grown substantially thanks to much empirical work.

\subsubsection{Available Models}

Most research in contemporary natural language processing makes use of a small number of neural architectures. Recurrent neural networks (RNNs) \citep{elman1990finding} such as LSTMs \citep{hochreiter1997long} and GRUs \citep{chung2014empirical} remain widely studied in LM probing, but Transformers \citep{vaswani2017attention} are dominant in modern NLP applications.

\subsubsection{The Inductive Biases of Neural Network Architectures}

Do any of these models represent a strict lower bound on humans' innate inductive bias? Strictly speaking, the answer is probably ``no''. It would be somewhat surprising if their inductive biases gave them absolutely no advantages over humans. These models have become widely used due to their empirical success in NLP applications. While this success is due in large part to the efficiency with which they can be trained on large corpora, it also suggests that they may have more advantageous inductive biases than other conceivable learners.

So what do we know about the inductive bias of these models, and are learnability results from them likely to generalize to humans? A growing body of work helps to address these questions by evaluating neural networks for a variety of human-like inductive bias. 

Numerous studies have found that ANNs lack a variety of human-like inductive biases prior to self-supervised training.
One striking example is that humans, but not ANNs, show a strong \emph{compositionality bias}. A key property of language is that words and phrases in language make stable compositional contributions to the semantics of larger constituents \citep{montague1973proper,fodor1988connectionism}. One consequence of this is that humans can understand the compositional semantic contribution of a newly learned word in any appropriate context \citep{lake2019human}. However, ANNs at human-like data scales have shown a general inability to make compositional generalizations \citep{lake2018generalization,kim2020cogs,keysers2020measuring}.

ANNs also generally lack a bias towards adopting hierarchical generalizations. \citet{mccoy2020does} test several varieties of RNNs without any pretraining using the Poverty of the Stimulus method on an ambiguous subject auxiliary inversion task, and find that none converge on a systematic hierarchical generalization. Subsequently, \citet{petty2021transformers} have shown a similar result for Transformers.

The fact that ANNs appear to lack these human-like biases might make them more appropriate model learners for two reasons. First, it means they probably do not have any special innate advantage over humans in these respects. Second, if the goal of the study is to establish, for example, whether an innate structural bias is necessary for learning some target, then an off-the-shelf ANN is already a relatively appropriate test subject without any special modification to remove the bias in question. However, much stronger evidence about their inductive biases is needed for strong existence proofs.


One practical question is whether there is an advantage to using RNNs or Transformers as model learners. RNNs have a strong locality bias \citep{dhingra2018neural,ravfogel2019studying} which Transformers lack. This is a consequence of the models' architectures: RNNs have the notion of linear order built in, since they get information about the rest of the sequence only from the previous token's output. Transformers on the other hand only receive information about linear order through a set of dedicated \newterm{positional embeddings} added to the input. As a result, Transformers must learn the semantics of positional embeddings including notions like locality from scratch.

On the other hand, the differences between the biases of LSTMs and Transformers may be weaker than one might expect when it comes to grammar learning. For example, \citet{warstadt2020blimp} compute the correlation between the accuracy scores of pairs of LMs on BLiMP. Among a population of models including an $n$-gram model, an LSTM, and two Transformers, they found that the most strongly correlated models were the LSTM and one of the Transformers.

\subsection{Summary}

While current widely available neural networks are far from ideal model learners, learnability results from them are likely to be relevant to human language acquisition. Most compellingly, a randomly initialized Transformer or LSTM has little innate bias in favor of valid linguistic generalizations. In fact, these models have generally weak inductive bias, as evidenced by their large hypothesis spaces and effectiveness in many domains besides language. Put another way, if we compare an arbitrarily selected human infant and an artificial neural network, both without any exposure to language, we should expect the human to place more prior probability on valid linguistic generalizations. This thought experiment provides some assurance that artificial neural networks, on the whole, do not possess substantial innate advantages over humans when it comes to language learning. 

Still, we cannot conclude with certainty that our models have no advantages at all. We are limited by our understanding of and our ability to quantify the inductive bias of humans and models, and models certainly show some superhuman or un-human-like abilities. Furthermore, while we have some ability to modify inductive biases of neural networks through architectural changes, we have nowhere near the same degree of control that we can exercise over the learning environment. This puts some practical limitations on---but does not totally rule out---the possibility of both improving the cognitive plausibility of our models and performing controlled experiments on model learners' inductive biases.

\section{Discussion}\label{sec:discussion}

We set out to determine what artificial neural networks can teach us about human language learning. We have shown that in the best-case scenario, model learners will be able to prove that specific linguistic behaviors are learnable under impoverished conditions, and thereby help to establish the causal roles of hypothesized advantages in the learning environment and the learner. We have also outlined the path leading up to this best-case scenario. However, we are still far from the best-case scenario. What does this mean for work that is already being done in this area?

\subsection{The Case for Model Learners}

At present, the strongest justifications for studying neural networks as models of human learners involve expense, ethics, and the potential for new experimental paradigms. Research on artificial learners is more scalable than research on children. With current language model architectures and hardware, a full simulation of the entire language acquisition period of a human learner takes on the order of one week on a single computer.\footnote{This estimate is based on a reproduction of the pretraining procedure used by \citet{warstadt2020learning}.}
Parallelization can make this even faster. Very little hands-on work is required during this training period.

Experimentation on artificial learners comes with few ethical restrictions. This is in contrast to experiments on human subjects---and especially infants---which must present minimal risk of harm to the subject. By design, ablations are often very harmful to learning outcomes, meaning we can never do an ablation on L1 acquisition in humans. Aside from experimentation on artificial language learning in humans, the only acquisition ablations we can do on language acquisition is in model learners. 

Finally, the use of model learners unlocks many novel experimental paradigms which are infeasible with human subjects for a number of reasons. With simulations, we have access to all aspects of the learning algorithm, the learner, and the learning environment. Machine learning methods provide many ways to manipulate the learning algorithm, offering choices from training objectives and regularization to curricula and multitask training. We can manipulate the learner's internal structure by simply changing neural architectures or hyperparameters such as depth, or making causal interventions such as changes to individual neurons \citep{vig2020investigating,finlayson2021causal} or interpretable linguistic features \citep{ravfogel2020null,elazar2021amnesic}. But arguably the greatest potential is in our ability to manipulate the learning environment. We are not limited to manipulating the size or source of the training data. Starting with a naturalistic corpus, we can manipulate the distribution of syntactic phenomena and word types, add noise, or inject counterfactual phenomena. 

\subsection{The Future of Model Learners}

One thing is clear: Machine learning and NLP are advancing at an unprecedented rate. This makes the prospect of using artificial learners as models of human language acquisition an especially salient possibility. In the last decade, there has been remarkable progress in the abilities of artificial learners to process human language and our ability to access these models. It is only natural that such a shift in our understanding of the learnability of language learning should have some real impact on debates about human language acquisition.

To deliver on this goal, we must make conscious choices to build more ecologically valid learners and learning environments. While the NLP community could make substantial progress on this problem, the focus is more often on improving the state of the art on well-known NLP tasks at whatever cost. Similarly, while language model probing and ``BERTology'' have become substantial subfields in recent years \citep{rogers2020primer}, this work often focuses on landmark models like BERT or the state of the art at the time. To obtain more useful cognitive models, cognitive scientists need to deliberately cultivate a model-building research agenda that builds on NLP, but with separate objectives. Benchmarks and competitions that set strict upper bounds on the quantity and nature of pretraining data could focus attention on this objective. Having a population of more plausible model learners will enable researchers to use existing LM probing methods to accelerate progress on questions in human language acquisition. 

While work on artificial learners in the near future is unlikely to yield incontrovertible proof about human learnability, we do not consider this cause for despair. A model learner that does not meet the stringent conditions of having no advantage over humans can still contribute converging evidence about human learnability. The evidence becomes stronger as we construct more plausible learning environments and learners.

\section*{Acknowledgments}

We are grateful to Ailis Cournane, Michael C.~Frank, Najoung Kim, Tal Linzen, Will Merrill, and Grusha Prasad for comments on this chapter. This work was improved by discussions with audiences at the Machine Learning for Language group at NYU, the Computational and Psycholinguistics Lab at NYU and Johns Hopkins, the Stanford Language and Cognition Lab, RyCoLab at ETH Zurich, and the Cognitive Machine Learning group at Ecole Normale Superieure, and with Roger Levy and Ellie Pavlick. We also thank the editors Shalom Lappin and Jean-Philippe Bernardy, and an anonymous reviewer for helpful comments.

This project has benefited from financial support to SB by Eric and Wendy Schmidt (made by recommendation of the Schmidt Futures program), Samsung Research (under the project \textit{Improving Deep Learning using Latent Structure}), Apple, and Intuit. This material is based upon work supported by the National Science Foundation under Grant No. 1850208. Any opinions, findings, and conclusions or recommendations expressed in this material are those of the author(s) and do not necessarily reflect the views of the National Science Foundation.

\bibliography{bibtex_example}

\begin{thebibliography}{166}
\providecommand{\natexlab}[1]{#1}
\providecommand{\url}[1]{\texttt{#1}}
\expandafter\ifx\csname urlstyle\endcsname\relax
  \providecommand{\doi}[1]{doi: #1}\else
  \providecommand{\doi}{doi: \begingroup \urlstyle{rm}\Url}\fi

\bibitem[Abend et~al.(2017)Abend, Kwiatkowski, Smith, Goldwater, and
  Steedman]{abend2017bootstrapping}
Omri Abend, Tom Kwiatkowski, Nathaniel~J. Smith, Sharon Goldwater, and Mark
  Steedman.
\newblock Bootstrapping language acquisition.
\newblock \emph{Cognition}, 164:\penalty0 116--143, July 2017.
\newblock ISSN 00100277.
\newblock \doi{10.1016/j.cognition.2017.02.009}.
\newblock URL
  \url{https://linkinghub.elsevier.com/retrieve/pii/S0010027717300495}.

\bibitem[Adi et~al.(2017)Adi, Kermany, Belinkov, Lavi, and
  Goldberg]{adi2017fine}
Yossi Adi, Einat Kermany, Yonatan Belinkov, Ofer Lavi, and Yoav Goldberg.
\newblock Fine-grained analysis of sentence embeddings using auxiliary
  prediction tasks.
\newblock In \emph{Proceedings of ICLR Conference Track. Toulon, France.},
  2017.

\bibitem[Austin(1962)]{austin1962how}
JL~Austin.
\newblock \emph{How to {Do} {Things} {With} {Words}}.
\newblock Oxford University Press, 1962.

\bibitem[Baker(1978)]{baker1978introduction}
Carl~Lee Baker.
\newblock \emph{Introduction to {Generative}-{Transformational} {Synta}}.
\newblock Prentice-Hall, Englewood Cliffs, NJ, 1978.

\bibitem[Baroni(2021)]{baroni2021proper}
Marco Baroni.
\newblock On the proper role of linguistically-oriented deep net analysis in
  linguistic theorizing.
\newblock \emph{arXiv:2106.08694 [cs]}, June 2021.
\newblock URL \url{http://arxiv.org/abs/2106.08694}.
\newblock arXiv: 2106.08694.

\bibitem[Belinkov and Glass(2019)]{belinkov2019analysis}
Yonatan Belinkov and James~R. Glass.
\newblock Analysis {Methods} in {Neural} {Language} {Processing}: {A} {Survey}.
\newblock \emph{Transactions of the Association for Computational Linguistics},
  7:\penalty0 49--72, 2019.
\newblock URL \url{https://www.aclweb.org/anthology/Q19-1004.pdf}.

\bibitem[Berwick et~al.(2011)Berwick, Pietroski, Yankama, and
  Chomsky]{berwick2011poverty}
Robert~C Berwick, Paul Pietroski, Beracah Yankama, and Noam Chomsky.
\newblock Poverty of the stimulus revisited.
\newblock \emph{Cognitive Science}, 35\penalty0 (7):\penalty0 1207--1242, 2011.

\bibitem[Bouchacourt and Baroni(2018)]{bouchacourt2018how}
Diane Bouchacourt and Marco Baroni.
\newblock How agents see things: {On} visual representations in an emergent
  language game.
\newblock In \emph{Proceedings of the 2018 conference on empirical methods in
  natural language processing}, pages 981--985, Brussels, Belgium, 2018.
  Association for Computational Linguistics.
\newblock \doi{10.18653/v1/D18-1119}.
\newblock URL \url{https://aclanthology.org/D18-1119}.

\bibitem[Brown et~al.(2020)Brown, Mann, Ryder, Subbiah, Kaplan, Dhariwal,
  Neelakantan, Shyam, Sastry, Askell, Agarwal, Herbert-Voss, Krueger, Henighan,
  Child, Ramesh, Ziegler, Wu, Winter, Hesse, Chen, Sigler, Litwin, Gray, Chess,
  Clark, Berner, McCandlish, Radford, Sutskever, and Amodei]{brown2020language}
Tom~B. Brown, Benjamin Mann, Nick Ryder, Melanie Subbiah, Jared Kaplan,
  Prafulla Dhariwal, Arvind Neelakantan, Pranav Shyam, Girish Sastry, Amanda
  Askell, Sandhini Agarwal, Ariel Herbert-Voss, Gretchen Krueger, Tom Henighan,
  Rewon Child, Aditya Ramesh, Daniel~M. Ziegler, Jeffrey Wu, Clemens Winter,
  Christopher Hesse, Mark Chen, Eric Sigler, Mateusz Litwin, Scott Gray,
  Benjamin Chess, Jack Clark, Christopher Berner, Sam McCandlish, Alec Radford,
  Ilya Sutskever, and Dario Amodei.
\newblock Language {Models} are {Few}-{Shot} {Learners}.
\newblock In \emph{Advances in {Neural} {Information} {Processing} {Systems}},
  2020.
\newblock URL
  \url{https://papers.nips.cc/paper/2020/file/1457c0d6bfcb4967418bfb8ac142f64a-Paper.pdf}.

\bibitem[Chaabouni et~al.(2019)Chaabouni, Kharitonov, Dupoux, and
  Baroni]{chaabouni2019anti}
Rahma Chaabouni, Eugene Kharitonov, Emmanuel Dupoux, and Marco Baroni.
\newblock Anti-efficient encoding in emergent communication.
\newblock \emph{Advances in Neural Information Processing Systems}, 32, 2019.

\bibitem[Chaabouni et~al.(2022)Chaabouni, Strub, Altché, Tarassov, Tallec,
  Davoodi, Mathewson, Tieleman, Lazaridou, and Piot]{chaabouni2022emergent}
Rahma Chaabouni, Florian Strub, Florent Altché, Eugene Tarassov, Corentin
  Tallec, Elnaz Davoodi, Kory~Wallace Mathewson, Olivier Tieleman, Angeliki
  Lazaridou, and Bilal Piot.
\newblock Emergent communication at scale.
\newblock In \emph{International conference on learning representations}, 2022.

\bibitem[Chang and Bergen(2022)]{chang2022word}
Tyler~A. Chang and Benjamin~K. Bergen.
\newblock Word {Acquisition} in {Neural} {Language} {Models}.
\newblock \emph{Transactions of the Association for Computational Linguistics},
  10:\penalty0 1--16, January 2022.
\newblock ISSN 2307-387X.
\newblock \doi{10.1162/tacl_a_00444}.
\newblock URL
  \url{https://direct.mit.edu/tacl/article/doi/10.1162/tacl_a_00444/109271/Word-Acquisition-in-Neural-Language-Models}.

\bibitem[Chaves(2020)]{chaves2020what}
Rui~P. Chaves.
\newblock What {Don}'t {RNN} {Language} {Models} {Learn} {About} {Filler}-{Gap}
  {Dependencies}?
\newblock In \emph{Proceedings of the third meeting of the {Society} for
  {Computation} in {Linguistics} ({SCiL})}, 2020.

\bibitem[Chen et~al.(2015)Chen, Fang, Lin, Vedantam, Gupta, Dollar, and
  Zitnick]{chen2015microsoft}
Xinlei Chen, Hao Fang, Tsung-Yi Lin, Ramakrishna Vedantam, Saurabh Gupta, Piotr
  Dollar, and C.~Lawrence Zitnick.
\newblock Microsoft {COCO} {Captions}: {Data} {Collection} and {Evaluation}
  {Server}.
\newblock In \emph{European conference on computer vision}, pages 740--755.
  Springer, April 2015.
\newblock URL \url{http://arxiv.org/abs/1504.00325}.
\newblock arXiv: 1504.00325.

\bibitem[Chen et~al.(2020)Chen, Li, Yu, El~Kholy, Ahmed, Gan, Cheng, and
  Liu]{chen2020uniter}
Yen-Chun Chen, Linjie Li, Licheng Yu, Ahmed El~Kholy, Faisal Ahmed, Zhe Gan,
  Yu~Cheng, and Jingjing Liu.
\newblock {UNITER}: {UNiversal} image-text representation learning.
\newblock In Andrea Vedaldi, Horst Bischof, Thomas Brox, and Jan-Michael Frahm,
  editors, \emph{Computer vision – {ECCV} 2020}, pages 104--120, Cham, 2020.
  Springer International Publishing.
\newblock ISBN 978-3-030-58577-8.

\bibitem[Chomsky(1965)]{chomsky1965aspects}
Noam Chomsky.
\newblock \emph{Aspects of the Theory of Syntax}.
\newblock MIT Press, 1965.

\bibitem[Chomsky(1971)]{chomsky1971problems}
Noam Chomsky.
\newblock Problems of knowledge and freedom: The {R}ussell lectures.
\newblock 1971.

\bibitem[Chomsky(2007)]{chomsky2007biolinguistic}
Noam Chomsky.
\newblock Biolinguistic {Explorations}: {Design}, {Development}, {Evolution}.
\newblock \emph{International Journal of Philosophical Studies}, 15\penalty0
  (1):\penalty0 1--21, January 2007.
\newblock ISSN 0967-2559, 1466-4542.
\newblock \doi{10.1080/09672550601143078}.
\newblock URL
  \url{http://www.tandfonline.com/doi/abs/10.1080/09672550601143078}.

\bibitem[Chomsky and Lasnik(1993)]{chomsky1993theory}
Noam Chomsky and Howard Lasnik.
\newblock The theory of principles and parameters.
\newblock In \emph{The minimalist program}. MIT Press, 1993.

\bibitem[Choshen et~al.(2021)Choshen, Hacohen, Weinshall, and
  Abend]{choshen2021grammar}
Leshem Choshen, Guy Hacohen, Daphna Weinshall, and Omri Abend.
\newblock The grammar-learning trajectories of neural language models.
\newblock \emph{arXiv preprint arXiv:2109.06096}, 2021.

\bibitem[Chouinard and Clark(2003)]{chouinard2003adult}
Michelle~M Chouinard and Eve~V Clark.
\newblock Adult reformulations of child errors as negative evidence.
\newblock \emph{Journal of child language}, 30\penalty0 (3):\penalty0 637--669,
  2003.
\newblock Publisher: Cambridge University Press.

\bibitem[Chowdhery et~al.(2022)Chowdhery, Narang, Devlin, Bosma, Mishra,
  Roberts, Barham, Chung, Sutton, Gehrmann, and {others}]{chowdhery2022palm}
Aakanksha Chowdhery, Sharan Narang, Jacob Devlin, Maarten Bosma, Gaurav Mishra,
  Adam Roberts, Paul Barham, Hyung~Won Chung, Charles Sutton, Sebastian
  Gehrmann, and {others}.
\newblock {PaLM}: {Scaling} language modeling with pathways.
\newblock \emph{arXiv preprint arXiv:2204.02311}, 2022.

\bibitem[Chowdhury and Zamparelli(2018)]{chowdhury2018rnn}
Shammur~Absar Chowdhury and Roberto Zamparelli.
\newblock {RNN} simulations of grammaticality judgments on long-distance
  dependencies.
\newblock In \emph{Proceedings of the 27th international conference on
  computational linguistics}, pages 133--144, 2018.

\bibitem[Christiansen and Chater(2016)]{christiansen2016creating}
Morten~H Christiansen and Nick Chater.
\newblock \emph{Creating language: Integrating evolution, acquisition, and
  processing}.
\newblock MIT Press, 2016.

\bibitem[Chung et~al.(2014)Chung, Gulcehre, Cho, and
  Bengio]{chung2014empirical}
Junyoung Chung, Caglar Gulcehre, KyungHyun Cho, and Yoshua Bengio.
\newblock Empirical {Evaluation} of {Gated} {Recurrent} {Neural} {Networks} on
  {Sequence} {Modeling}.
\newblock \emph{arXiv:1412.3555 [cs]}, December 2014.
\newblock URL \url{http://arxiv.org/abs/1412.3555}.
\newblock arXiv: 1412.3555.

\bibitem[Clark and Lappin(2011)]{clark2011linguistic}
Alexander Clark and Shalom Lappin.
\newblock \emph{Linguistic {Nativism} and the {Poverty} of the {Stimulus}}.
\newblock John Wiley \& Sons, 2011.

\bibitem[Coulton(1972)]{coulton1972princes}
G.~G. Coulton.
\newblock The {Princes} of the {World}.
\newblock In \emph{From {St}. {Francis} to {Dante}}, Translations from the
  {Chronicle} of the {Franciscan} {Salimbene}, 1221-1288, pages 239--256.
  University of Pennsylvania Press, 2 edition, 1972.
\newblock ISBN 978-0-8122-7672-5.
\newblock URL \url{https://www.jstor.org/stable/j.ctv4t8279.25}.

\bibitem[Crain and Nakayama(1987)]{crain1987structure}
Stephen Crain and Mineharu Nakayama.
\newblock Structure {Dependence} in {Grammar} {Formation}.
\newblock \emph{Language}, 63\penalty0 (3):\penalty0 522--543, 1987.
\newblock URL \url{https://www.jstor.org/stable/415004}.

\bibitem[Cristia et~al.(2019)Cristia, Dupoux, Gurven, and
  Stieglitz]{cristia2019child}
Alejandrina Cristia, Emmanuel Dupoux, Michael Gurven, and Jonathan Stieglitz.
\newblock Child‐{Directed} {Speech} {Is} {Infrequent} in a
  {Forager}‐{Farmer} {Population}: {A} {Time} {Allocation} {Study}.
\newblock \emph{Child Development}, 90\penalty0 (3):\penalty0 759--773, May
  2019.
\newblock ISSN 0009-3920, 1467-8624.
\newblock \doi{10.1111/cdev.12974}.
\newblock URL \url{https://onlinelibrary.wiley.com/doi/10.1111/cdev.12974}.

\bibitem[Dai and Le(2015)]{dai2015semisupervised}
Andrew~M Dai and Quoc~V Le.
\newblock Semi-supervised sequence learning.
\newblock In C.~Cortes, N.~Lawrence, D.~Lee, M.~Sugiyama, and R.~Garnett,
  editors, \emph{Advances in neural information processing systems}, volume~28.
  Curran Associates, Inc., 2015.
\newblock URL
  \url{https://proceedings.neurips.cc/paper/2015/file/7137debd45ae4d0ab9aa953017286b20-Paper.pdf}.

\bibitem[Davies(2009)]{davies2009385}
Mark Davies.
\newblock The 385+ million word {Corpus} of {Contemporary} {American} {English}
  (1990–2008+): {Design}, architecture, and linguistic insights.
\newblock \emph{International journal of corpus linguistics}, 14\penalty0
  (2):\penalty0 159--190, 2009.
\newblock Publisher: John Benjamins.

\bibitem[Devlin et~al.(2019)Devlin, Chang, Lee, and Toutanova]{devlin2019bert}
Jacob Devlin, Ming-Wei Chang, Kenton Lee, and Kristina Toutanova.
\newblock {BERT}: {Pre}-training of {Deep} {Bidirectional} {Transformers} for
  {Language} {Understanding}.
\newblock In \emph{Proceedings of the 2019 {Conference} of the {North}
  {American} {Chapter} of the {Association} for {Computational} {Linguistics}:
  {Human} {Language} {Technologies}, {Volume} 1 ({Long} and {Short} {Papers})},
  pages 4171--4186, 2019.
\newblock URL \url{https://www.aclweb.org/anthology/N19-1423}.

\bibitem[Dhingra et~al.(2018)Dhingra, Jin, Yang, Cohen, and
  Salakhutdinov]{dhingra2018neural}
Bhuwan Dhingra, Qiao Jin, Zhilin Yang, William Cohen, and Ruslan Salakhutdinov.
\newblock Neural {Models} for {Reasoning} over {Multiple} {Mentions} {Using}
  {Coreference}.
\newblock In \emph{Proceedings of the 2018 {Conference} of the {North}
  {American} {Chapter} of the {Association} for {Computational} {Linguistics}:
  {Human} {Language} {Technologies}, {Volume} 2 ({Short} {Papers})}, pages
  42--48, 2018.

\bibitem[Dupoux(2018)]{dupoux2018cognitive}
Emmanuel Dupoux.
\newblock Cognitive {Science} in the era of {Artificial} {Intelligence}: {A}
  roadmap for reverse-engineering the infant language-learner.
\newblock \emph{Cognition}, 173:\penalty0 43--59, April 2018.
\newblock ISSN 00100277.
\newblock \doi{10.1016/j.cognition.2017.11.008}.
\newblock URL \url{http://arxiv.org/abs/1607.08723}.
\newblock arXiv: 1607.08723.

\bibitem[Dupre(2021)]{dupre2021what}
Gabe Dupre.
\newblock ({What}) {Can} {Deep} {Learning} {Contribute} to {Theoretical}
  {Linguistics}?
\newblock \emph{Minds and Machines}, September 2021.
\newblock ISSN 0924-6495, 1572-8641.
\newblock \doi{10.1007/s11023-021-09571-w}.
\newblock URL \url{https://link.springer.com/10.1007/s11023-021-09571-w}.

\bibitem[Elazar et~al.(2021{\natexlab{a}})Elazar, Kassner, Ravfogel,
  Ravichander, Hovy, Sch{\"u}tze, and Goldberg]{elazar2021measuring}
Yanai Elazar, Nora Kassner, Shauli Ravfogel, Abhilasha Ravichander, Eduard
  Hovy, Hinrich Sch{\"u}tze, and Yoav Goldberg.
\newblock Measuring and improving consistency in pretrained language models.
\newblock \emph{Transactions of the Association for Computational Linguistics},
  9:\penalty0 1012--1031, 2021{\natexlab{a}}.

\bibitem[Elazar et~al.(2021{\natexlab{b}})Elazar, Ravfogel, Jacovi, and
  Goldberg]{elazar2021amnesic}
Yanai Elazar, Shauli Ravfogel, Alon Jacovi, and Yoav Goldberg.
\newblock Amnesic {Probing}: {Behavioral} {Explanation} with {Amnesic}
  {Counterfactuals}.
\newblock \emph{Transactions of the Association for Computational Linguistics},
  9:\penalty0 160--175, 2021{\natexlab{b}}.

\bibitem[Elman(1990)]{elman1990finding}
Jeffrey~L Elman.
\newblock Finding structure in time.
\newblock \emph{Cognitive science}, 14\penalty0 (2):\penalty0 179--211, 1990.
\newblock Publisher: Wiley Online Library.

\bibitem[Ettinger et~al.(2016)Ettinger, Elgohary, and
  Resnik]{ettinger2016probing}
Allyson Ettinger, Ahmed Elgohary, and Philip Resnik.
\newblock Probing for semantic evidence of composition by means of simple
  classification tasks.
\newblock In \emph{Proceedings of the 1st {Workshop} on {Evaluating}
  {Vector}-{Space} {Representations} for {NLP}}, pages 134--139, 2016.
\newblock URL \url{https://www.aclweb.org/anthology/W16-2524}.

\bibitem[Finlayson et~al.(2021)Finlayson, Mueller, Gehrmann, Shieber, Linzen,
  and Belinkov]{finlayson2021causal}
Matthew Finlayson, Aaron Mueller, Sebastian Gehrmann, Stuart Shieber, Tal
  Linzen, and Yonatan Belinkov.
\newblock Causal {Analysis} of {Syntactic} {Agreement} {Mechanisms} in {Neural}
  {Language} {Models}.
\newblock In \emph{Proceedings of the 59th {Annual} {Meeting} of the
  {Association} for {Computational} {Linguistics} and the 11th {International}
  {Joint} {Conference} on {Natural} {Language} {Processing} ({Volume} 1: {Long}
  {Papers})}, pages 1828--1843, Online, 2021. Association for Computational
  Linguistics.
\newblock \doi{10.18653/v1/2021.acl-long.144}.
\newblock URL \url{https://aclanthology.org/2021.acl-long.144}.

\bibitem[Fodor and Crowther(2002)]{fodor2002understanding}
Janet~Dean Fodor and Carrie Crowther.
\newblock Understanding stimulus poverty arguments.
\newblock \emph{The Linguistic Review}, 18\penalty0 (1-2), January 2002.
\newblock ISSN 0167-6318, 1613-3676.
\newblock \doi{10.1515/tlir.19.1-2.105}.
\newblock URL
  \url{https://www.degruyter.com/document/doi/10.1515/tlir.19.1-2.105/html}.

\bibitem[Fodor and Pylyshyn(1988)]{fodor1988connectionism}
Jerry~A Fodor and Zenon~W Pylyshyn.
\newblock Connectionism and cognitive architecture: {A} critical analysis.
\newblock \emph{Cognition}, 28\penalty0 (1-2):\penalty0 3--71, 1988.
\newblock Publisher: Elsevier.

\bibitem[Frank et~al.(2017)Frank, Braginsky, Yurovsky, and
  Marchman]{frank2017wordbank}
Michael~C. Frank, Mika Braginsky, Daniel Yurovsky, and Virginia~A. Marchman.
\newblock Wordbank: {A}n open repository for developmental vocabulary data.
\newblock \emph{Journal of Child Language}, 44\penalty0 (3):\penalty0 677--694,
  May 2017.
\newblock ISSN 0305-0009, 1469-7602.
\newblock \doi{10.1017/S0305000916000209}.
\newblock URL
  \url{https://www.cambridge.org/core/product/identifier/S0305000916000209/type/journal_article}.

\bibitem[Frank and Mathis(2007)]{frank2007transformational}
Robert Frank and Donald Mathis.
\newblock Transformational networks.
\newblock \emph{Models of Human Language Acquisition}, page~22, 2007.

\bibitem[Fromkin et~al.(1974)Fromkin, Krashen, Curtiss, Rigler, and
  Rigler]{fromkin1974development}
Victoria Fromkin, Stephen Krashen, Susan Curtiss, David Rigler, and Marilyn
  Rigler.
\newblock The {Development} of {Language} in {Genie}: a {Case} of {Language}
  {Acquisition} beyond the "{Critical} {Period}".
\newblock \emph{Brain and Language}, 1:\penalty0 81--107, 1974.

\bibitem[Futrell et~al.(2021)Futrell, Gibson, Tily, Blank, Vishnevetsky,
  Piantadosi, and Fedorenko]{futrell2021natural}
Richard Futrell, Edward Gibson, Harry~J. Tily, Idan Blank, Anastasia
  Vishnevetsky, Steven~T. Piantadosi, and Evelina Fedorenko.
\newblock The {Natural} {Stories} corpus: a reading-time corpus of {English}
  texts containing rare syntactic constructions.
\newblock \emph{Language Resources and Evaluation}, 55\penalty0 (1):\penalty0
  63--77, March 2021.
\newblock ISSN 1574-020X, 1574-0218.
\newblock \doi{10.1007/s10579-020-09503-7}.
\newblock URL \url{https://link.springer.com/10.1007/s10579-020-09503-7}.

\bibitem[Gauthier et~al.(2020)Gauthier, Hu, Wilcox, Qian, and
  Levy]{gauthier2020syntaxgym}
Jon Gauthier, Jennifer Hu, Ethan Wilcox, Peng Qian, and Roger Levy.
\newblock {SyntaxGym}: {An} {Online} {Platform} for {Targeted} {Evaluation} of
  {Language} {Models}.
\newblock In \emph{Proceedings of the 58th {Annual} {Meeting} of the
  {Association} for {Computational} {Linguistics}: {System} {Demonstrations}},
  pages 70--76, Online, July 2020. Association for Computational Linguistics.
\newblock \doi{10.18653/v1/2020.acl-demos.10}.
\newblock URL \url{https://aclanthology.org/2020.acl-demos.10}.

\bibitem[Gentner(1982)]{gentner1982why}
Dedre Gentner.
\newblock Why nouns are learned before verbs: {Linguistic} relativity versus
  natural partitioning.
\newblock In S~Kuczaj, editor, \emph{Language development: {Language} cognition
  and culture.}, page~48. 1982.

\bibitem[Gilkerson et~al.(2017)Gilkerson, Richards, Warren, Montgomery,
  Greenwood, Kimbrough~Oller, Hansen, and Paul]{gilkerson2017mapping}
Jill Gilkerson, Jeffrey~A Richards, Steven~F Warren, Judith~K Montgomery,
  Charles~R Greenwood, D~Kimbrough~Oller, John~HL Hansen, and Terrance~D Paul.
\newblock Mapping the early language environment using all-day recordings and
  automated analysis.
\newblock \emph{American journal of speech-language pathology}, 26\penalty0
  (2):\penalty0 248--265, 2017.
\newblock Publisher: ASHA.

\bibitem[Gleitman and Wanner(1982)]{gleitman1982language}
Lila Gleitman and Eric Wanner.
\newblock Language {Acquisition}: {The} {State} of the {Art}.
\newblock In Lila Gleitman and Eric Wanner, editors, \emph{Language
  {Acquisition}: {The} {State} of the {Art}}. Cambridge University Press, 1982.

\bibitem[Gold(1967)]{gold1967language}
E~Mark Gold.
\newblock Language identification in the limit.
\newblock \emph{Information and control}, 10\penalty0 (5):\penalty0 447--474,
  1967.
\newblock Publisher: Elsevier.

\bibitem[Gordon(1985)]{gordon1985levelordering}
Peter Gordon.
\newblock Level-ordering in lexical development.
\newblock \emph{Cognition}, pages 73--93, 1985.

\bibitem[Gulordava et~al.(2019)Gulordava, Bojanowski, Grave, Linzen, and
  Baroni]{gulordava2019colorless}
Kristina Gulordava, Piotr Bojanowski, Edouard Grave, Tal Linzen, and Marco
  Baroni.
\newblock Colorless green recurrent networks dream hierarchically.
\newblock \emph{Proceedings of the Society for Computation in Linguistics},
  2\penalty0 (1):\penalty0 363--364, 2019.

\bibitem[Gómez and Gerken(2000)]{gomez2000infant}
Rebecca~L Gómez and LouAnn Gerken.
\newblock Infant artificial language learning and language acquisition.
\newblock \emph{Trends in cognitive sciences}, 4\penalty0 (5):\penalty0
  178--186, 2000.
\newblock Publisher: Elsevier.

\bibitem[Hale(2001)]{hale2001probabilistic}
John Hale.
\newblock A {Probabilistic} {Earley} {Parser} as a {Psycholinguistic} {Model}.
\newblock In \emph{Second {Meeting} of the {North} {American} {Chapter} of the
  {Association} for {Computational} {Linguistics}}, 2001.
\newblock URL \url{https://aclanthology.org/N01-1021}.

\bibitem[Han et~al.(2016)Han, Musolino, and Lidz]{han2016endogenous}
Chung-hye Han, Julien Musolino, and Jeffrey Lidz.
\newblock Endogenous sources of variation in language acquisition.
\newblock \emph{Proceedings of the National Academy of Sciences}, 113\penalty0
  (4):\penalty0 942--947, January 2016.
\newblock ISSN 0027-8424, 1091-6490.
\newblock \doi{10.1073/pnas.1517094113}.
\newblock URL \url{https://pnas.org/doi/full/10.1073/pnas.1517094113}.

\bibitem[Hart and Risley(1992)]{hart1992american}
Betty Hart and Todd~R. Risley.
\newblock American parenting of language-learning children: {Persisting}
  differences in family-child interactions observed in natural home
  environments.
\newblock \emph{Developmental Psychology}, 28\penalty0 (6):\penalty0 1096,
  1992.
\newblock URL \url{https://psycnet.apa.org/fulltext/1993-09151-001.pdf}.
\newblock Publisher: American Psychological Association.

\bibitem[Haussler(1990)]{haussler1990probably}
David Haussler.
\newblock Probably approximately correct learning.
\newblock In \emph{Proceedings of the eighth national conference on artificial
  intelligence}, pages 1101--1108. AAAI Press, 1990.

\bibitem[He et~al.(2020)He, Liu, Gao, and Chen]{he2020deberta}
Pengcheng He, Xiaodong Liu, Jianfeng Gao, and Weizhu Chen.
\newblock {DeBERTa}: {Decoding}-enhanced {BERT} with {Disentangled}
  {Attention}.
\newblock In \emph{International conference on learning representations}, 2020.

\bibitem[Hewitt and Liang(2019)]{hewitt2019designing}
John Hewitt and Percy Liang.
\newblock Designing and {Interpreting} {Probes} with {Control} {Tasks}.
\newblock In \emph{Conference on {Empirical} {Methods} in {Natural} {Language}
  {Processing}}. Association for Computational Linguistics, 2019.
\newblock URL \url{https://www.aclweb.org/anthology/D19-1275}.
\newblock event-place: Hong Kong.

\bibitem[Hewitt and Manning(2019)]{hewitt2019structural}
John Hewitt and Christopher~D Manning.
\newblock A structural probe for finding syntax in word representations.
\newblock In \emph{Proceedings of the 2019 Conference of the North American
  Chapter of the Association for Computational Linguistics: Human Language
  Technologies, Volume 1 (Long and Short Papers)}, pages 4129--4138, 2019.

\bibitem[Hochreiter and Schmidhuber(1997)]{hochreiter1997long}
Sepp Hochreiter and Jürgen Schmidhuber.
\newblock Long short-term memory.
\newblock \emph{Neural Computation}, 9\penalty0 (8):\penalty0 1735--1780, 1997.
\newblock Publisher: MIT Press.

\bibitem[Howard and Ruder(2018)]{howard2018universal}
Jeremy Howard and Sebastian Ruder.
\newblock Universal {Language} {Model} {Fine}-tuning for {Text}
  {Classification}.
\newblock In \emph{Proceedings of the 56th {Annual} {Meeting} of the
  {Association} for {Computational} {Linguistics} ({Volume} 1: {Long}
  {Papers})}, pages 328--339, 2018.

\bibitem[Howell et~al.(2005)Howell, Jankowicz, and Becker]{howell2005model}
Steve~R. Howell, Damian Jankowicz, and Suzanna Becker.
\newblock A model of grounded language acquisition: {Sensorimotor} features
  improve lexical and grammatical learning.
\newblock \emph{Journal of Memory and Language}, 53\penalty0 (2):\penalty0
  258--276, August 2005.
\newblock ISSN 0749596X.
\newblock \doi{10.1016/j.jml.2005.03.002}.
\newblock URL
  \url{https://linkinghub.elsevier.com/retrieve/pii/S0749596X05000495}.

\bibitem[Hu et~al.(2020)Hu, Gauthier, Qian, Wilcox, and Levy]{hu2020systematic}
Jennifer Hu, Jon Gauthier, Peng Qian, Ethan Wilcox, and Roger Levy.
\newblock A {Systematic} {Assessment} of {Syntactic} {Generalization} in
  {Neural} {Language} {Models}.
\newblock In \emph{Proceedings of the 58th {Annual} {Meeting} of the
  {Association} for {Computational} {Linguistics}}, pages 1725--1744, Online,
  July 2020. Association for Computational Linguistics.
\newblock URL \url{https://www.aclweb.org/anthology/2020.acl-main.158}.

\bibitem[Huang et~al.(2019)Huang, Le~Bras, Bhagavatula, and
  Choi]{huang-etal-2019-cosmos}
Lifu Huang, Ronan Le~Bras, Chandra Bhagavatula, and Yejin Choi.
\newblock Cosmos {QA}: {Machine} reading comprehension with contextual
  commonsense reasoning.
\newblock In \emph{Proceedings of the 2019 conference on empirical methods in
  natural language processing and the 9th international joint conference on
  natural language processing ({EMNLP}-{IJCNLP})}, pages 2391--2401, Hong Kong,
  China, November 2019. Association for Computational Linguistics.
\newblock \doi{10.18653/v1/D19-1243}.
\newblock URL \url{https://aclanthology.org/D19-1243}.

\bibitem[Huebner and Willits(2021)]{huebner2021using}
Philip~A. Huebner and Jon~A. Willits.
\newblock Using lexical context to discover the noun category: {Younger}
  children have it easier.
\newblock In Kara~D. Federmeier and Lili Sahakyan, editors, \emph{The context
  of cognition: {Emerging} perspectives}, volume~75 of \emph{Psychology of
  learning and motivation}, pages 279--331. Academic Press, 2021.
\newblock \doi{https://doi.org/10.1016/bs.plm.2021.08.002}.
\newblock URL
  \url{https://www.sciencedirect.com/science/article/pii/S0079742121000256}.
\newblock ISSN: 0079-7421.

\bibitem[Iki and Aizawa(2021)]{iki2021effecta}
Taichi Iki and Akiko Aizawa.
\newblock Effect of {Visual} {Extensions} on {Natural} {Language}
  {Understanding} in {Vision}-and-{Language} {Models}.
\newblock In \emph{Proceedings of the 2021 {Conference} on {Empirical}
  {Methods} in {Natural} {Language} {Processing}}, pages 2189--2196, Online and
  Punta Cana, Dominican Republic, 2021. Association for Computational
  Linguistics.
\newblock \doi{10.18653/v1/2021.emnlp-main.167}.
\newblock URL \url{https://aclanthology.org/2021.emnlp-main.167}.

\bibitem[Kamath et~al.(2021)Kamath, Singh, LeCun, Synnaeve, Misra, and
  Carion]{kamath2021mdetr}
Aishwarya Kamath, Mannat Singh, Yann LeCun, Gabriel Synnaeve, Ishan Misra, and
  Nicolas Carion.
\newblock {MDETR}-modulated detection for end-to-end multi-modal understanding.
\newblock In \emph{Proceedings of the IEEE/CVF International Conference on
  Computer Vision}, pages 1780--1790, 2021.

\bibitem[Katz(1987)]{katz1987estimation}
Slava Katz.
\newblock Estimation of probabilities from sparse data for the language model
  component of a speech recognizer.
\newblock \emph{IEEE transactions on acoustics, speech, and signal processing},
  35\penalty0 (3):\penalty0 400--401, 1987.

\bibitem[Keysers et~al.(2020)Keysers, Schärli, Scales, Buisman, Furrer,
  Kashubin, Momchev, Sinopalnikov, Stafiniak, Tihon, Tsarkov, Wang, van Zee,
  and Bousquet]{keysers2020measuring}
Daniel Keysers, Nathanael Schärli, Nathan Scales, Hylke Buisman, Daniel
  Furrer, Sergii Kashubin, Nikola Momchev, Danila Sinopalnikov, Lukasz
  Stafiniak, Tibor Tihon, Dmitry Tsarkov, Xiao Wang, Marc van Zee, and Olivier
  Bousquet.
\newblock Measuring {Compositional} {Generalization}: {A} {Comprehensive}
  {Method} on {Realistic} {Data}.
\newblock \emph{arXiv:1912.09713 [cs, stat]}, June 2020.
\newblock URL \url{http://arxiv.org/abs/1912.09713}.
\newblock arXiv: 1912.09713.

\bibitem[Kim and Linzen(2020)]{kim2020cogs}
Najoung Kim and Tal Linzen.
\newblock {COGS}: {A} {Compositional} {Generalization} {Challenge} {Based} on
  {Semantic} {Interpretation}.
\newblock In \emph{Proceedings of the 2020 {Conference} on {Empirical}
  {Methods} in {Natural} {Language} {Processing} ({EMNLP})}, pages 9087--9105,
  Online, November 2020. Association for Computational Linguistics.
\newblock \doi{10.18653/v1/2020.emnlp-main.731}.
\newblock URL \url{https://aclanthology.org/2020.emnlp-main.731}.

\bibitem[Kimball(1973)]{kimball1973formal}
John~P. Kimball.
\newblock \emph{The {Formal} {Theory} of {Grammar}}.
\newblock Prentice-Hall, Englewood Cliffs, NJ, 1973.

\bibitem[Kirby(1999)]{kirby1999function}
Simon Kirby.
\newblock \emph{Function, selection, and innateness: {The} emergence of
  language universals}.
\newblock OUP Oxford, 1999.

\bibitem[Kottur et~al.(2017)Kottur, Moura, Lee, and
  Batra]{kottur-etal-2017-natural}
Satwik Kottur, José Moura, Stefan Lee, and Dhruv Batra.
\newblock Natural language does not emerge `naturally' in multi-agent dialog.
\newblock In \emph{Proceedings of the 2017 conference on empirical methods in
  natural language processing}, pages 2962--2967, Copenhagen, Denmark,
  September 2017. Association for Computational Linguistics.
\newblock \doi{10.18653/v1/D17-1321}.
\newblock URL \url{https://aclanthology.org/D17-1321}.

\bibitem[Krishna et~al.(2017)Krishna, Zhu, Groth, Johnson, Hata, Kravitz, Chen,
  Kalantidis, Li, Shamma, Bernstein, and Fei-Fei]{krishna2017visual}
Ranjay Krishna, Yuke Zhu, Oliver Groth, Justin Johnson, Kenji Hata, Joshua
  Kravitz, Stephanie Chen, Yannis Kalantidis, Li-Jia Li, David~A. Shamma,
  Michael~S. Bernstein, and Li~Fei-Fei.
\newblock Visual {Genome}: {Connecting} {Language} and {Vision} {Using}
  {Crowdsourced} {Dense} {Image} {Annotations}.
\newblock \emph{International Journal of Computer Vision}, 123\penalty0
  (1):\penalty0 32--73, May 2017.
\newblock ISSN 0920-5691, 1573-1405.
\newblock \doi{10.1007/s11263-016-0981-7}.
\newblock URL \url{http://link.springer.com/10.1007/s11263-016-0981-7}.

\bibitem[Lake and Baroni(2018)]{lake2018generalization}
Brenden~M Lake and Marco Baroni.
\newblock Generalization without systematicity: {On} the compositional skills
  of sequence-to-sequence recurrent networks.
\newblock In \emph{International {Conference} on {Machine} {Learning}}, pages
  2879--2888, 2018.

\bibitem[Lake et~al.(2019)Lake, Linzen, and Baroni]{lake2019human}
Brenden~M Lake, Tal Linzen, and Marco Baroni.
\newblock Human few-shot learning of compositional instructions.
\newblock In \emph{Proceedings of the 41st {Annual} {Conference} of the
  {Cognitive} {Science} {Society}}, May 2019.

\bibitem[Lakhotia et~al.(2021)Lakhotia, Kharitonov, Hsu, Adi, Polyak, Bolte,
  Nguyen, Copet, Baevski, Mohamed, and Dupoux]{lakhotia2021generative}
Kushal Lakhotia, Evgeny Kharitonov, Wei-Ning Hsu, Yossi Adi, Adam Polyak,
  Benjamin Bolte, Tu-Anh Nguyen, Jade Copet, Alexei Baevski, Adelrahman
  Mohamed, and Emmanuel Dupoux.
\newblock Generative {Spoken} {Language} {Modeling} from {Raw} {Audio}.
\newblock \emph{arXiv:2102.01192 [cs]}, September 2021.
\newblock URL \url{http://arxiv.org/abs/2102.01192}.
\newblock arXiv: 2102.01192.

\bibitem[Landauer and Dutnais(1997)]{landauer1997solution}
Thomas~K Landauer and Susan~T Dutnais.
\newblock A {Solution} to {Plato}'s {Problem}: {The} {Latent} {Semantic}
  {Analysis} {Theory} of {Acquisition}, {Induction}, and {Representation} of
  {Knowledge}.
\newblock \emph{Psychological Review}, 104\penalty0 (2):\penalty0 211--240,
  1997.

\bibitem[Lau et~al.(2017)Lau, Clark, and Lappin]{lau2017grammaticality}
Jey~Han Lau, Alexander Clark, and Shalom Lappin.
\newblock Grammaticality, acceptability, and probability: {A} probabilistic
  view of linguistic knowledge.
\newblock \emph{Cognitive Science}, 41\penalty0 (5):\penalty0 1202--1241, 2017.
\newblock Publisher: Wiley Online Library.

\bibitem[Lavechin et~al.(2022)Lavechin, de~Seyssel, Métais, Metze, Mohamed,
  BREDIN, Dupoux, and Cristia]{lavechin2022early}
Marvin Lavechin, Maureen de~Seyssel, Marianne Métais, Florian Metze,
  Abdelrahman Mohamed, Hervé BREDIN, Emmanuel Dupoux, and Alejandrina Cristia.
\newblock Early phonetic learning from ecological audio: domain-general versus
  domain-specific mechanisms.
\newblock 2022.
\newblock Publisher: PsyArXiv.

\bibitem[Lazaridou and Baroni(2020)]{lazaridou2020emergent}
Angeliki Lazaridou and Marco Baroni.
\newblock Emergent {Multi}-{Agent} {Communication} in the {Deep} {Learning}
  {Era}.
\newblock \emph{arXiv:2006.02419 [cs]}, July 2020.
\newblock URL \url{http://arxiv.org/abs/2006.02419}.
\newblock arXiv: 2006.02419.

\bibitem[Lazaridou et~al.(2015)Lazaridou, Pham, and
  Baroni]{lazaridou2015combining}
Angeliki Lazaridou, Nghia~The Pham, and Marco Baroni.
\newblock Combining {Language} and {Vision} with a {Multimodal} {Skip}-gram
  {Model}.
\newblock In \emph{Proceedings of the 2015 {Conference} of the {North}
  {American} {Chapter} of the {Association} for {Computational} {Linguistics}:
  {Human} {Language} {Technologies}}, pages 153--163, Denver, Colorado, May
  2015. Association for Computational Linguistics.
\newblock \doi{10.3115/v1/N15-1016}.
\newblock URL \url{https://aclanthology.org/N15-1016}.

\bibitem[Lazaridou et~al.(2017)Lazaridou, Peysakhovich, and
  Baroni]{lazaridou2017multiagent}
Angeliki Lazaridou, Alexander Peysakhovich, and Marco Baroni.
\newblock Multi-{Agent} {Cooperation} and the {Emergence} of ({Natural})
  {Language}.
\newblock In \emph{International {Conference} on {Learning} {Representations}},
  March 2017.
\newblock URL \url{http://arxiv.org/abs/1612.07182}.
\newblock arXiv: 1612.07182.

\bibitem[Lazaridou et~al.(2018)Lazaridou, Hermann, Tuyls, and
  Clark]{lazaridou2018emergence}
Angeliki Lazaridou, Karl~Moritz Hermann, Karl Tuyls, and Stephen Clark.
\newblock Emergence of linguistic communication from referential games with
  symbolic and pixel input.
\newblock In \emph{International conference on learning representations}, 2018.

\bibitem[Lazaridou et~al.(2020)Lazaridou, Potapenko, and
  Tieleman]{lazaridou2020multiagent}
Angeliki Lazaridou, Anna Potapenko, and Olivier Tieleman.
\newblock Multi-agent {Communication} meets {Natural} {Language}: {Synergies}
  between {Functional} and {Structural} {Language} {Learning}.
\newblock In \emph{Proceedings of the 58th {Annual} {Meeting} of the
  {Association} for {Computational} {Linguistics}}, pages 7663--7674, Online,
  July 2020. Association for Computational Linguistics.
\newblock \doi{10.18653/v1/2020.acl-main.685}.
\newblock URL \url{https://aclanthology.org/2020.acl-main.685}.

\bibitem[LeCun et~al.(2015)LeCun, Bengio, and Hinton]{lecun2015deep}
Yann LeCun, Yoshua Bengio, and Geoffrey Hinton.
\newblock Deep learning.
\newblock \emph{Nature}, 521\penalty0 (7553):\penalty0 436, 2015.

\bibitem[Legate and Yang(2002)]{legate2002empirical}
Julie~Anne Legate and Charles~D Yang.
\newblock Empirical re-assessment of stimulus poverty arguments.
\newblock \emph{The Linguistic Review}, 18\penalty0 (1-2):\penalty0 151--162,
  2002.
\newblock Publisher: Walter de Gruyter.

\bibitem[Lerdahl et~al.(1983)Lerdahl, Jackendoff, and
  Jackendoff]{lerdahl1983generative}
Fred Lerdahl, Ray~S Jackendoff, and Ray Jackendoff.
\newblock \emph{A Generative Theory of Tonal Music}.
\newblock MIT Press, 1983.

\bibitem[Levy(2008)]{levy2008expectationbased}
Roger Levy.
\newblock Expectation-based syntactic comprehension.
\newblock \emph{Cognition}, 106\penalty0 (3):\penalty0 1126--1177, March 2008.
\newblock ISSN 00100277.
\newblock \doi{10.1016/j.cognition.2007.05.006}.
\newblock URL
  \url{https://linkinghub.elsevier.com/retrieve/pii/S0010027707001436}.

\bibitem[Li and Bowling(2019)]{li2019ease}
Fushan Li and Michael Bowling.
\newblock Ease-of-teaching and language structure from emergent communication.
\newblock \emph{Advances in neural information processing systems}, 32, 2019.

\bibitem[Lidz et~al.(2003)Lidz, Waxman, and Freedman]{lidz2003what}
Jeffrey Lidz, Sandra Waxman, and Jennifer Freedman.
\newblock What infants know about syntax but couldn't have learned:
  {E}xperimental evidence for syntactic structure at 18 months.
\newblock \emph{Cognition}, 89\penalty0 (3):\penalty0 295--303, October 2003.
\newblock ISSN 00100277.
\newblock \doi{10.1016/S0010-0277(03)00116-1}.
\newblock URL
  \url{https://linkinghub.elsevier.com/retrieve/pii/S0010027703001161}.

\bibitem[Linzen(2019)]{linzen2019can}
Tal Linzen.
\newblock What can linguistics and deep learning contribute to each other?
  {Response} to {Pater}.
\newblock \emph{Language}, 95\penalty0 (1):\penalty0 e99--e108, 2019.
\newblock Publisher: Linguistic Society of America.

\bibitem[Linzen and Baroni(2021)]{linzen2021syntactic}
Tal Linzen and Marco Baroni.
\newblock Syntactic {Structure} from {Deep} {Learning}.
\newblock \emph{Annual Reviews of Linguistics}, 2021.

\bibitem[Linzen et~al.(2016)Linzen, Dupoux, and Goldberg]{linzen2016assessing}
Tal Linzen, Emmanuel Dupoux, and Yoav Goldberg.
\newblock Assessing the ability of {LSTM}s to learn syntax-sensitive
  dependencies.
\newblock \emph{Transactions of the Association for Computational Linguistics},
  4:\penalty0 521--535, 2016.

\bibitem[Lison and Tiedemann(2016)]{lison2016opensubtitles2016}
Pierre Lison and Jorg Tiedemann.
\newblock {OpenSubtitles2016}: {Extracting} {Large} {Parallel} {Corpora} from
  {Movie} and {TV} {Subtitles}.
\newblock In \emph{Proceedings of the 10th {International} {Conference} on
  {Language} {Resources} and {Evaluation} ({LREC} 2016)}, page~7, 2016.

\bibitem[Liu et~al.(2019)Liu, Ott, Goyal, Du, Joshi, Chen, Levy, Lewis,
  Zettlemoyer, and Stoyanov]{liu2019roberta}
Yinhan Liu, Myle Ott, Naman Goyal, Jingfei Du, Mandar Joshi, Danqi Chen, Omer
  Levy, Mike Lewis, Luke Zettlemoyer, and Veselin Stoyanov.
\newblock {RoBERTa}: {A} robustly optimized {BERT} pretraining approach.
\newblock \emph{arXiv preprint arXiv:1907.11692}, 2019.
\newblock URL \url{http://arxiv.org/abs/1907.11692}.

\bibitem[Lovering et~al.(2021)Lovering, Jha, Linzen, and
  Pavlick]{lovering2021predicting}
Charles Lovering, Rohan Jha, Tal Linzen, and Ellie Pavlick.
\newblock Predicting {Inductive} {Biases} of {Fine}-tuned {Models}.
\newblock In \emph{International {Conference} on {Learning} {Representations}},
  2021.
\newblock URL \url{https://openreview.net/forum?id=mNtmhaDkAr}.

\bibitem[Lu et~al.(2019)Lu, Batra, Parikh, and Lee]{lu2019vilbert}
Jiasen Lu, Dhruv Batra, Devi Parikh, and Stefan Lee.
\newblock {ViLBERT}: {Pretraining} {Task}-{Agnostic} {Visiolinguistic}
  {Representations} for {Vision}-and-{Language} {Tasks}.
\newblock In \emph{Advances in {Neural} {Information} {Processing} {Systems}},
  volume~32. Curran Associates, Inc., 2019.
\newblock URL
  \url{https://proceedings.neurips.cc/paper/2019/hash/c74d97b01eae257e44aa9d5bade97baf-Abstract.html}.

\bibitem[MacWhinney(2014)]{macwhinney2014childes}
Brian MacWhinney.
\newblock \emph{The {CHILDES} project: {Tools} for analyzing talk, {Volume}
  {II}: {The} database.}
\newblock Psychology Press, 2014.

\bibitem[Mankewitz et~al.()Mankewitz, Boyce, Waldon, Loukatou, Yu, Mu, Goodman,
  and Frank]{mankewitz2021multi}
Jessica Mankewitz, Veronica Boyce, Brandon Waldon, Georgia Loukatou, Dhara Yu,
  Jesse Mu, Noah~D Goodman, and Michael~C Frank.
\newblock Multi-party referential communication in complex strategic games.
\newblock Publisher: PsyArXiv.

\bibitem[Manning(2015)]{manning2015computational}
Christopher~D. Manning.
\newblock Computational {Linguistics} and {Deep} {Learning}.
\newblock \emph{Computational Linguistics}, 41\penalty0 (4):\penalty0 701--707,
  December 2015.
\newblock ISSN 0891-2017, 1530-9312.
\newblock \doi{10.1162/COLI_a_00239}.
\newblock URL \url{https://direct.mit.edu/coli/article/41/4/701-707/1512}.

\bibitem[Manning et~al.(2020)Manning, Clark, Hewitt, Khandelwal, and
  Levy]{manning2020emergent}
Christopher~D. Manning, Kevin Clark, John Hewitt, Urvashi Khandelwal, and Omer
  Levy.
\newblock Emergent linguistic structure in artificial neural networks trained
  by self-supervision.
\newblock \emph{Proceedings of the National Academy of Sciences}, 117\penalty0
  (48):\penalty0 30046--30054, December 2020.
\newblock ISSN 0027-8424, 1091-6490.
\newblock \doi{10.1073/pnas.1907367117}.
\newblock URL \url{http://www.pnas.org/lookup/doi/10.1073/pnas.1907367117}.

\bibitem[Marcus(1993)]{marcus1993negative}
Gary~F Marcus.
\newblock Negative evidence in language acquisition.
\newblock \emph{Cognition}, 46\penalty0 (1):\penalty0 53--85, 1993.
\newblock Publisher: Elsevier.

\bibitem[Marvin and Linzen(2018)]{marvin2018targeted}
Rebecca Marvin and Tal Linzen.
\newblock Targeted {Syntactic} {Evaluation} of {Language} {Models}.
\newblock In \emph{Proceedings of the 2018 {Conference} on {Empirical}
  {Methods} in {Natural} {Language} {Processing}}, pages 1192--1202, 2018.

\bibitem[McCoy et~al.(2018)McCoy, Frank, and Linzen]{mccoy2018revisiting}
R~Thomas McCoy, Robert Frank, and Tal Linzen.
\newblock Revisiting the poverty of the stimulus: hierarchical generalization
  without a hierarchical bias in recurrent neural networks.
\newblock In \emph{Proceedings of the 40th {Annual} {Conference} of the
  {Cognitive} {Science} {Society}.}, 2018.

\bibitem[McCoy et~al.(2020)McCoy, Frank, and Linzen]{mccoy2020does}
R.~Thomas McCoy, Robert Frank, and Tal Linzen.
\newblock Does {Syntax} {Need} to {Grow} on {Trees}? {Sources} of
  {Hierarchical} {Inductive} {Bias} in {Sequence}-to-{Sequence} {Networks}.
\newblock \emph{Transactions of the Association for Computational Linguistics},
  8:\penalty0 125--140, December 2020.
\newblock ISSN 2307-387X.
\newblock \doi{10.1162/tacl_a_00304}.
\newblock URL \url{https://direct.mit.edu/tacl/article/43542}.

\bibitem[Mitchell(1980)]{mitchell1980need}
Tom~M Mitchell.
\newblock \emph{The need for biases in learning generalizations}.
\newblock Department of Computer Science, Laboratory for Computer Science
  Research …, 1980.

\bibitem[Montague(1973)]{montague1973proper}
Richard Montague.
\newblock The proper treatment of quantification in ordinary {English}.
\newblock In \emph{Approaches to natural language}, pages 221--242. Springer,
  1973.

\bibitem[Nguyen et~al.(2020)Nguyen, de~Seyssel, Rozé, Rivière, Kharitonov,
  Baevski, Dunbar, and Dupoux]{nguyen2020zero}
Tu~Anh Nguyen, Maureen de~Seyssel, Patricia Rozé, Morgane Rivière, Evgeny
  Kharitonov, Alexei Baevski, Ewan Dunbar, and Emmanuel Dupoux.
\newblock The {Zero} {Resource} {Speech} {Benchmark} 2021: {Metrics} and
  baselines for unsupervised spoken language modeling.
\newblock \emph{arXiv:2011.11588 [cs, eess]}, December 2020.
\newblock URL \url{http://arxiv.org/abs/2011.11588}.
\newblock arXiv: 2011.11588.

\bibitem[Orhan et~al.()Orhan, Gupta, and Lake]{orhanselfsupervised}
A~Emin Orhan, Vaibhav~V Gupta, and Brenden~M Lake.
\newblock Self-supervised learning through the eyes of a child.
\newblock page~12.

\bibitem[Pannitto and Herbelot(2020)]{pannitto2020recurrent}
Ludovica Pannitto and Aurélie Herbelot.
\newblock Recurrent babbling: evaluating the acquisition of grammar from
  limited input data.
\newblock In \emph{Proceedings of the 24th {Conference} on {Computational}
  {Natural} {Language} {Learning}}, pages 165--176, Online, November 2020.
  Association for Computational Linguistics.
\newblock \doi{10.18653/v1/2020.conll-1.13}.
\newblock URL \url{https://aclanthology.org/2020.conll-1.13}.

\bibitem[Papadimitriou et~al.(2021)Papadimitriou, Chi, Futrell, and
  Mahowald]{papadimitriou2021deep}
Isabel Papadimitriou, Ethan~A. Chi, Richard Futrell, and Kyle Mahowald.
\newblock Deep subjecthood: {Higher}-order grammatical features in multilingual
  {BERT}.
\newblock In \emph{Proceedings of the 16th conference of the european chapter
  of the association for computational linguistics: {Main} volume}, pages
  2522--2532, Online, April 2021. Association for Computational Linguistics.
\newblock \doi{10.18653/v1/2021.eacl-main.215}.
\newblock URL \url{https://aclanthology.org/2021.eacl-main.215}.

\bibitem[Pater(2019)]{pater2019generative}
Joe Pater.
\newblock Generative linguistics and neural networks at 60: {Foundation},
  friction, and fusion.
\newblock \emph{Language}, 95\penalty0 (1):\penalty0 e41--e74, 2019.
\newblock Publisher: Linguistic Society of America.

\bibitem[Perfors et~al.(2011)Perfors, Tenenbaum, and
  Regier]{perfors2011learnability}
Andy Perfors, Joshua~B Tenenbaum, and Terry Regier.
\newblock The learnability of abstract syntactic principles.
\newblock \emph{Cognition}, 118\penalty0 (3):\penalty0 306--338, 2011.
\newblock Publisher: Elsevier.

\bibitem[Peters et~al.(2018)Peters, Neumann, Iyyer, Gardner, Clark, Lee, and
  Zettlemoyer]{peters2018}
Matthew Peters, Mark Neumann, Mohit Iyyer, Matt Gardner, Christopher Clark,
  Kenton Lee, and Luke Zettlemoyer.
\newblock Deep contextualized word representations.
\newblock In \emph{Proceedings of the 2018 Conference of the North American
  Chapter of the Association for Computational Linguistics: Human Language
  Technologies, Volume 1 (Long Papers)}, pages 2227--2237. Association for
  Computational Linguistics, 2018.
\newblock \doi{10.18653/v1/N18-1202}.
\newblock URL \url{http://aclweb.org/anthology/N18-1202}.

\bibitem[Petroni et~al.(2019)Petroni, Rocktäschel, Riedel, Lewis, Bakhtin, Wu,
  and Miller]{petroni2019language}
Fabio Petroni, Tim Rocktäschel, Sebastian Riedel, Patrick Lewis, Anton
  Bakhtin, Yuxiang Wu, and Alexander Miller.
\newblock Language {Models} as {Knowledge} {Bases}?
\newblock In \emph{Proceedings of the 2019 {Conference} on {Empirical}
  {Methods} in {Natural} {Language} {Processing} and the 9th {International}
  {Joint} {Conference} on {Natural} {Language} {Processing}
  ({EMNLP}-{IJCNLP})}, pages 2463--2473, Hong Kong, China, November 2019.
  Association for Computational Linguistics.
\newblock \doi{10.18653/v1/D19-1250}.
\newblock URL \url{https://www.aclweb.org/anthology/D19-1250}.

\bibitem[Petty and Frank(2021)]{petty2021transformers}
Jackson Petty and Robert Frank.
\newblock Transformers {Generalize} {Linearly}.
\newblock \emph{arXiv:2109.12036 [cs]}, September 2021.
\newblock URL \url{http://arxiv.org/abs/2109.12036}.
\newblock arXiv: 2109.12036.

\bibitem[Pimentel et~al.(2020)Pimentel, Valvoda, Hall~Maudslay, Zmigrod,
  Williams, and Cotterell]{pimentel2020informationtheoretic}
Tiago Pimentel, Josef Valvoda, Rowan Hall~Maudslay, Ran Zmigrod, Adina
  Williams, and Ryan Cotterell.
\newblock Information-{Theoretic} {Probing} for {Linguistic} {Structure}.
\newblock In \emph{Proceedings of the 58th {Annual} {Meeting} of the
  {Association} for {Computational} {Linguistics}}, pages 4609--4622, Online,
  July 2020. Association for Computational Linguistics.
\newblock \doi{10.18653/v1/2020.acl-main.420}.
\newblock URL \url{https://www.aclweb.org/anthology/2020.acl-main.420}.

\bibitem[Pérez-Mayos et~al.(2021)Pérez-Mayos, Ballesteros, and
  Wanner]{perez-mayos2021how}
Laura Pérez-Mayos, Miguel Ballesteros, and Leo Wanner.
\newblock How much pretraining data do language models need to learn syntax?
\newblock September 2021.
\newblock URL \url{https://arxiv.org/abs/2109.03160v2}.

\bibitem[Radford et~al.(2018)Radford, Narasimhan, Salimans, and
  Sutskever]{radford2018improving}
Alec Radford, Karthik Narasimhan, Tim Salimans, and Ilya Sutskever.
\newblock Improving language understanding with unsupervised learning.
\newblock Technical report, Technical report, OpenAI, 2018.

\bibitem[Radford et~al.(2021)Radford, Kim, Hallacy, Ramesh, Goh, Agarwal,
  Sastry, Askell, Mishkin, Clark, and {others}]{radford2021learning}
Alec Radford, Jong~Wook Kim, Chris Hallacy, Aditya Ramesh, Gabriel Goh,
  Sandhini Agarwal, Girish Sastry, Amanda Askell, Pamela Mishkin, Jack Clark,
  and {others}.
\newblock Learning transferable visual models from natural language
  supervision.
\newblock \emph{arXiv preprint arXiv:2103.00020}, 2021.

\bibitem[Rae et~al.(2021)Rae, Borgeaud, Cai, Millican, Hoffmann, Song,
  Aslanides, Henderson, Ring, Young, and {others}]{rae2021scaling}
Jack~W Rae, Sebastian Borgeaud, Trevor Cai, Katie Millican, Jordan Hoffmann,
  Francis Song, John Aslanides, Sarah Henderson, Roman Ring, Susannah Young,
  and {others}.
\newblock Scaling language models: {Methods}, analysis \& insights from
  training gopher.
\newblock \emph{arXiv preprint arXiv:2112.11446}, 2021.

\bibitem[Rasin and Aravind(2021)]{rasin2021nature}
Ezer Rasin and Athulya Aravind.
\newblock The nature of the semantic stimulus: the acquisition of every as a
  case study.
\newblock \emph{Natural Language Semantics}, 29\penalty0 (2):\penalty0
  339--375, June 2021.
\newblock ISSN 0925-854X, 1572-865X.
\newblock \doi{10.1007/s11050-020-09168-6}.
\newblock URL \url{https://link.springer.com/10.1007/s11050-020-09168-6}.

\bibitem[Ravfogel et~al.(2019)Ravfogel, Goldberg, and
  Linzen]{ravfogel2019studying}
Shauli Ravfogel, Yoav Goldberg, and Tal Linzen.
\newblock Studying the {Inductive} {Biases} of {RNNs} with {Synthetic}
  {Variations} of {Natural} {Languages}.
\newblock In \emph{Proceedings of {NAACL}-{HLT}}, pages 3532--3542, 2019.

\bibitem[Ravfogel et~al.(2020)Ravfogel, Elazar, Gonen, Twiton, and
  Goldberg]{ravfogel2020null}
Shauli Ravfogel, Yanai Elazar, Hila Gonen, Michael Twiton, and Yoav Goldberg.
\newblock Null {It} {Out}: {Guarding} {Protected} {Attributes} by {Iterative}
  {Nullspace} {Projection}.
\newblock In \emph{Proceedings of the 58th {Annual} {Meeting} of the
  {Association} for {Computational} {Linguistics}}, pages 7237--7256, Online,
  2020. Association for Computational Linguistics.
\newblock \doi{10.18653/v1/2020.acl-main.647}.
\newblock URL \url{https://www.aclweb.org/anthology/2020.acl-main.647}.

\bibitem[Reali and Christiansen(2005)]{reali2005uncovering}
Florencia Reali and Morten~H Christiansen.
\newblock Uncovering the richness of the stimulus: {Structure} dependence and
  indirect statistical evidence.
\newblock \emph{Cognitive Science}, 29\penalty0 (6):\penalty0 1007--1028, 2005.
\newblock Publisher: Wiley Online Library.

\bibitem[Ren et~al.(2019)Ren, Guo, Labeau, Cohen, and
  Kirby]{ren2019compositional}
Yi~Ren, Shangmin Guo, Matthieu Labeau, Shay~B Cohen, and Simon Kirby.
\newblock Compositional languages emerge in a neural iterated learning model.
\newblock In \emph{International conference on learning representations}, 2019.

\bibitem[Resnick et~al.(2020)Resnick, Gupta, Foerster, Dai, and
  Cho]{resnick2020capacity}
Cinjon Resnick, Abhinav Gupta, Jakob Foerster, Andrew~M Dai, and Kyunghyun Cho.
\newblock Capacity, bandwidth, and compositionality in emergent language
  learning.
\newblock In \emph{Proceedings of the 19th international conference on
  autonomous agents and {MultiAgent} systems}, pages 1125--1133, 2020.

\bibitem[Reuland(2017)]{reuland2017grammar}
Eric Reuland.
\newblock Grammar of binding in the languages of the world: {Unity} versus
  diversity.
\newblock \emph{Cognition}, 168:\penalty0 370--379, November 2017.
\newblock ISSN 00100277.
\newblock \doi{10.1016/j.cognition.2016.01.020}.
\newblock URL
  \url{https://linkinghub.elsevier.com/retrieve/pii/S0010027716300208}.

\bibitem[Rita et~al.(2020)Rita, Chaabouni, and Dupoux]{rita2020lazimpa}
Mathieu Rita, Rahma Chaabouni, and Emmanuel Dupoux.
\newblock “{LazImpa}”: {Lazy} and {Impatient} neural agents learn to
  communicate efficiently.
\newblock In \emph{Proceedings of the 24th conference on computational natural
  language learning}, pages 335--343, Online, November 2020. Association for
  Computational Linguistics.
\newblock URL \url{https://www.aclweb.org/anthology/2020.conll-1.26}.

\bibitem[Rogers et~al.(2020)Rogers, Kovaleva, and Rumshisky]{rogers2020primer}
Anna Rogers, Olga Kovaleva, and Anna Rumshisky.
\newblock A {Primer} in {BERTology}: {What} we know about how {BERT} works.
\newblock In \emph{Findings of {EMNLP}}, 2020.
\newblock URL \url{https://www.aclweb.org/anthology/2020.tacl-1.54}.

\bibitem[Sakaguchi et~al.(2020)Sakaguchi, Bras, Bhagavatula, and
  Choi]{sakaguchi2020winogrande}
Keisuke Sakaguchi, Ronan~Le Bras, Chandra Bhagavatula, and Yejin Choi.
\newblock {WinoGrande}: {An} {Adversarial} {Winograd} {Schema} {Challenge} at
  {Scale}.
\newblock In \emph{Proceedings of the {AAAI} {Conference} on {Artificial}
  {Intelligence}}, volume 34(05), pages 8732--8740, 2020.
\newblock \doi{https://doi.org/10.1609/aaai.v34i05.6399}.

\bibitem[Salazar et~al.(2020)Salazar, Liang, Nguyen, and
  Kirchhoff]{salazar2020masked}
Julian Salazar, Davis Liang, Toan~Q. Nguyen, and Katrin Kirchhoff.
\newblock Masked {Language} {Model} {Scoring}.
\newblock In \emph{Proceedings of the 58th {Annual} {Meeting} of the
  {Association} for {Computational} {Linguistics}}, pages 2699--2712, Online,
  July 2020. Association for Computational Linguistics.
\newblock \doi{10.18653/v1/2020.acl-main.240}.
\newblock URL \url{https://www.aclweb.org/anthology/2020.acl-main.240}.

\bibitem[Sch{\"u}tze(1996)]{schutze1996empirical}
Carson~T. Sch{\"u}tze.
\newblock \emph{The Empirical Base of Linguistics: Grammaticality Judgments and
  Linguistic Methodology}.
\newblock University of Chicago Press, 1996.

\bibitem[Searle(1969)]{searle1969speech}
John~R. Searle.
\newblock \emph{Speech acts: {An} essay in the philosophy of language}, volume
  626.
\newblock Cambridge University Press, 1969.

\bibitem[Shi et~al.(2016)Shi, Padhi, and Knight]{shi2016syntax}
Xing Shi, Inkit Padhi, and Kevin Knight.
\newblock Does string-based neural {MT} learn source syntax?
\newblock In \emph{Proceedings of the 2016 Conference on Empirical Methods in
  Natural Language Processing}, pages 1526--1534, 2016.

\bibitem[Singh et~al.(2021)Singh, Hu, Goswami, Couairon, Galuba, Rohrbach, and
  Kiela]{singh2021flava}
Amanpreet Singh, Ronghang Hu, Vedanuj Goswami, Guillaume Couairon, Wojciech
  Galuba, Marcus Rohrbach, and Douwe Kiela.
\newblock Flava: A foundational language and vision alignment model.
\newblock \emph{arXiv preprint arXiv:2112.04482}, 2021.

\bibitem[Soderstrom et~al.(2003)Soderstrom, Seidl, Nelson, and
  Jusczyk]{soderstrom2003prosodic}
Melanie Soderstrom, Amanda Seidl, Deborah G~Kemler Nelson, and Peter~W Jusczyk.
\newblock The prosodic bootstrapping of phrases: {Evidence} from prelinguistic
  infants.
\newblock \emph{Journal of Memory and Language}, 49\penalty0 (2):\penalty0
  249--267, 2003.
\newblock Publisher: Elsevier.

\bibitem[Sprouse et~al.(2013)Sprouse, Sch{\"u}tze, and Almeida]{sprouse2013li}
Jon Sprouse, Carson~T. Sch{\"u}tze, and Diogo Almeida.
\newblock A comparison of informal and formal acceptability judgments using a
  random sample from {L}inguistic {I}nquiry 2001--2010.
\newblock \emph{Lingua}, 134:\penalty0 219--248, 2013.

\bibitem[Su et~al.(2020)Su, Zhu, Cao, Li, Lu, Wei, and Dai]{Su2020VL-BERT:}
Weijie Su, Xizhou Zhu, Yue Cao, Bin Li, Lewei Lu, Furu Wei, and Jifeng Dai.
\newblock Vl-bert: Pre-training of generic visual-linguistic representations.
\newblock In \emph{International Conference on Learning Representations}, 2020.
\newblock URL \url{https://openreview.net/forum?id=SygXPaEYvH}.

\bibitem[Sullivan et~al.(2021)Sullivan, Mei, Perfors, Wojcik, and
  Frank]{sullivan2021saycam}
Jessica Sullivan, Michelle Mei, Andrew Perfors, Erica Wojcik, and Michael~C.
  Frank.
\newblock {SAYCam}: {A} {Large}, {Longitudinal} {Audiovisual} {Dataset}
  {Recorded} {From} the {Infant}’s {Perspective}.
\newblock \emph{Open Mind}, 5:\penalty0 20--29, May 2021.
\newblock ISSN 2470-2986.
\newblock \doi{10.1162/opmi_a_00039}.
\newblock URL
  \url{https://direct.mit.edu/opmi/article/doi/10.1162/opmi_a_00039/97495/SAYCam-A-Large-Longitudinal-Audiovisual-Dataset}.

\bibitem[Sun et~al.(2019)Sun, Myers, Vondrick, Murphy, and
  Schmid]{sun2019videobert}
Chen Sun, Austin Myers, Carl Vondrick, Kevin Murphy, and Cordelia Schmid.
\newblock Videobert: {A} joint model for video and language representation
  learning.
\newblock In \emph{Proceedings of the {IEEE}/{CVF} international conference on
  computer vision}, pages 7464--7473, 2019.

\bibitem[Tan and Bansal(2019)]{tan2019lxmert}
Hao Tan and Mohit Bansal.
\newblock {LXMERT}: {Learning} {Cross}-{Modality} {Encoder} {Representations}
  from {Transformers}.
\newblock In \emph{Proceedings of the 2019 {Conference} on {Empirical}
  {Methods} in {Natural} {Language} {Processing} and the 9th {International}
  {Joint} {Conference} on {Natural} {Language} {Processing}
  ({EMNLP}-{IJCNLP})}, pages 5100--5111, Hong Kong, China, November 2019.
  Association for Computational Linguistics.
\newblock \doi{10.18653/v1/D19-1514}.
\newblock URL \url{https://aclanthology.org/D19-1514}.

\bibitem[Tenney et~al.(2019)Tenney, Xia, Chen, Wang, Poliak, McCoy, Kim,
  Van~Durme, Bowman, Das, et~al.]{tenney2019you}
Ian Tenney, Patrick Xia, Berlin Chen, Alex Wang, Adam Poliak, R~Thomas McCoy,
  Najoung Kim, Benjamin Van~Durme, Samuel~R Bowman, Dipanjan Das, et~al.
\newblock What do you learn from context? probing for sentence structure in
  contextualized word representations.
\newblock In \emph{Proceedings of ICLR}, 2019.

\bibitem[Valiant(1984)]{valiant1984theory}
L.~G. Valiant.
\newblock A theory of the learnable.
\newblock \emph{Communications of the ACM}, 27\penalty0 (11):\penalty0
  1134--1142, November 1984.
\newblock ISSN 0001-0782, 1557-7317.
\newblock \doi{10.1145/1968.1972}.
\newblock URL \url{https://dl.acm.org/doi/10.1145/1968.1972}.

\bibitem[Van~Dooren et~al.()Van~Dooren, Dieuleveut, Cournane, and
  Hacquard]{vandoorenAcceptedfiguring}
Annmarie Van~Dooren, Anouk Dieuleveut, Ail\'{i}s Cournane, and Valentine
  Hacquard.
\newblock Figuring out root and epistemic uses for modals: {The} role of the
  input.
\newblock \emph{Journal of Semantics}.

\bibitem[van Schijndel et~al.(2019)van Schijndel, Mueller, and
  Linzen]{vanschijndel2019quantity}
Marten van Schijndel, Aaron Mueller, and Tal Linzen.
\newblock Quantity doesn't buy quality syntax with neural language models.
\newblock In \emph{Proceedings of the 2019 {Conference} on {Empirical}
  {Methods} in {Natural} {Language} {Processing} and the 9th {International}
  {Joint} {Conference} on {Natural} {Language} {Processing}
  ({EMNLP}-{IJCNLP})}, pages 5831--5837, Hong Kong, China, November 2019.
  Association for Computational Linguistics.
\newblock \doi{10.18653/v1/D19-1592}.
\newblock URL \url{https://www.aclweb.org/anthology/D19-1592}.

\bibitem[Vaswani et~al.(2017)Vaswani, Shazeer, Parmar, Uszkoreit, Jones, Gomez,
  Kaiser, and Polosukhin]{vaswani2017attention}
Ashish Vaswani, Noam Shazeer, Niki Parmar, Jakob Uszkoreit, Llion Jones,
  Aidan~N Gomez, {\textbackslash}Lukasz Kaiser, and Illia Polosukhin.
\newblock Attention is {All} you {Need}.
\newblock In I.~Guyon, U.~V. Luxburg, S.~Bengio, H.~Wallach, R.~Fergus,
  S.~Vishwanathan, and R.~Garnett, editors, \emph{Advances in {Neural}
  {Information} {Processing} {Systems} 30}, pages 5998--6008. Curran
  Associates, Inc., 2017.
\newblock URL
  \url{http://papers.nips.cc/paper/7181-attention-is-all-you-need.pdf}.

\bibitem[Vig et~al.(2020)Vig, Gehrmann, Belinkov, Qian, Nevo, Singer, and
  Shieber]{vig2020investigating}
Jesse Vig, Sebastian Gehrmann, Yonatan Belinkov, Sharon Qian, Daniel Nevo,
  Yaron Singer, and Stuart Shieber.
\newblock Investigating gender bias in language models using causal mediation
  analysis.
\newblock In H.~Larochelle, M.~Ranzato, R.~Hadsell, M.~F. Balcan, and H.~Lin,
  editors, \emph{Advances in neural information processing systems}, volume~33,
  pages 12388--12401. Curran Associates, Inc., 2020.
\newblock URL
  \url{https://proceedings.neurips.cc/paper/2020/file/92650b2e92217715fe312e6fa7b90d82-Paper.pdf}.

\bibitem[Voita and Titov(2020)]{voita2020informationtheoretic}
Elena Voita and Ivan Titov.
\newblock Information-{Theoretic} {Probing} with {Minimum} {Description}
  {Length}.
\newblock In \emph{Proceedings of the 2020 {Conference} on {Empirical}
  {Methods} in {Natural} {Language} {Processing}}, Punta Cana, Dominican
  Republic, November 2020. Association for Computational Linguistics.
\newblock URL \url{https://www.aclweb.org/anthology/2020.emnlp-main.14.pdf}.

\bibitem[Wang and Cho(2019)]{wang2019bert}
Alex Wang and Kyunghyun Cho.
\newblock {BERT} has a mouth, and it must speak: {BERT} as a {Markov} random
  field language model.
\newblock In \emph{Proceedings of the workshop on methods for optimizing and
  evaluating neural language generation}, pages 30--36, Minneapolis, Minnesota,
  June 2019. Association for Computational Linguistics.
\newblock \doi{10.18653/v1/W19-2304}.
\newblock URL \url{https://aclanthology.org/W19-2304}.

\bibitem[Warstadt and Bowman(2019)]{warstadt2019linguistic}
Alex Warstadt and Samuel~R. Bowman.
\newblock Linguistic {Analysis} of {Pretrained} {Sentence} {Encoders} with
  {Acceptability} {Judgments}.
\newblock \emph{arXiv preprint arXiv:1901.03438}, 2019.

\bibitem[Warstadt and Bowman(2020)]{warstadt2020can}
Alex Warstadt and Samuel~R Bowman.
\newblock Can neural networks acquire a structural bias from raw linguistic
  data?
\newblock In \emph{Proceedings of the 42nd {Annual} {Conference} of the
  {Cognitive} {Science} {Society}.}, 2020.

\bibitem[Warstadt et~al.(2018)Warstadt, Singh, and Bowman]{warstadt2018neural}
Alex Warstadt, Amanpreet Singh, and Samuel~R Bowman.
\newblock Neural network acceptability judgments.
\newblock \emph{arXiv preprint arXiv:1805.12471}, 2018.

\bibitem[Warstadt et~al.(2020{\natexlab{a}})Warstadt, Parrish, Liu, Mohananey,
  Peng, Wang, and Bowman]{warstadt2020blimp}
Alex Warstadt, Alicia Parrish, Haokun Liu, Anhad Mohananey, Wei Peng, Sheng-Fu
  Wang, and Samuel~R. Bowman.
\newblock {BLiMP}: {The} {Benchmark} of {Linguistic} {Minimal} {Pairs} for
  {English}.
\newblock \emph{Transactions of the Association for Computational Linguistics},
  8:\penalty0 377--392, 2020{\natexlab{a}}.
\newblock \doi{10.1162/tacl_a_00321}.
\newblock URL \url{https://doi.org/10.1162/tacl_a_00321}.

\bibitem[Warstadt et~al.(2020{\natexlab{b}})Warstadt, Zhang, Li, Liu, and
  Bowman]{warstadt2020learning}
Alex Warstadt, Yian Zhang, Haau-Sing Li, Haokun Liu, and Samuel~R Bowman.
\newblock Learning {Which} {Features} {Matter}: {RoBERTa} {Acquires} a
  {Preference} for {Linguistic} {Generalizations} ({Eventually}).
\newblock In \emph{Proceedings of the 2020 {Conference} on {Empirical}
  {Methods} in {Natural} {Language} {Processing}}, Punta Cana, Dominican
  Republic, November 2020{\natexlab{b}}. Association for Computational
  Linguistics.

\bibitem[Wilcox et~al.(2018)Wilcox, Levy, Morita, and Futrell]{wilcox2018what}
Ethan Wilcox, Roger Levy, Takashi Morita, and Richard Futrell.
\newblock What do {RNN} {Language} {Models} {Learn} about {Filler}–{Gap}
  {Dependencies}?
\newblock In \emph{Proceedings of the 2018 {EMNLP} {Workshop} {BlackboxNLP}:
  {Analyzing} and {Interpreting} {Neural} {Networks} for {NLP}}, pages
  211--221, 2018.

\bibitem[Wilcox et~al.(2021)Wilcox, Vani, and Levy]{wilcox2021targeted}
Ethan Wilcox, Pranali Vani, and Roger Levy.
\newblock A {Targeted} {Assessment} of {Incremental} {Processing} in {Neural}
  {Language} {Models} and {Humans}.
\newblock In \emph{Proceedings of the 59th {Annual} {Meeting} of the
  {Association} for {Computational} {Linguistics} and the 11th {International}
  {Joint} {Conference} on {Natural} {Language} {Processing} ({Volume} 1: {Long}
  {Papers})}, pages 939--952, Online, August 2021. Association for
  Computational Linguistics.
\newblock \doi{10.18653/v1/2021.acl-long.76}.
\newblock URL \url{https://aclanthology.org/2021.acl-long.76}.

\bibitem[Wilson and Izmailov(2020)]{wilson2020bayesian}
Andrew~G Wilson and Pavel Izmailov.
\newblock Bayesian deep learning and a probabilistic perspective of
  generalization.
\newblock \emph{Advances in neural information processing systems},
  33:\penalty0 4697--4708, 2020.

\bibitem[Wilson(2006)]{wilson2006learning}
Colin Wilson.
\newblock Learning phonology with substantive bias: {An} experimental and
  computational study of velar palatalization.
\newblock \emph{Cognitive science}, 30\penalty0 (5):\penalty0 945--982, 2006.
\newblock Publisher: Wiley Online Library.

\bibitem[Yang and Piantadosi(2022)]{yang2022one}
Yuan Yang and Steven~T. Piantadosi.
\newblock One model for the learning of language.
\newblock \emph{Proceedings of the National Academy of Sciences}, 119\penalty0
  (5):\penalty0 e2021865119, February 2022.
\newblock ISSN 0027-8424, 1091-6490.
\newblock \doi{10.1073/pnas.2021865119}.
\newblock URL \url{https://pnas.org/doi/full/10.1073/pnas.2021865119}.

\bibitem[Yun et~al.(2021)Yun, Sun, and Pavlick]{yun2021does}
Tian Yun, Chen Sun, and Ellie Pavlick.
\newblock Does {Vision}-and-{Language} {Pretraining} {Improve} {Lexical}
  {Grounding}?
\newblock In \emph{Proceedings of {EMNLP}}, September 2021.
\newblock arXiv: 2109.10246.

\bibitem[Zellers et~al.(2018)Zellers, Bisk, Schwartz, and
  Choi]{zellers-etal-2018-swag}
Rowan Zellers, Yonatan Bisk, Roy Schwartz, and Yejin Choi.
\newblock {SWAG}: {A} large-scale adversarial dataset for grounded commonsense
  inference.
\newblock In \emph{Proceedings of the 2018 conference on empirical methods in
  natural language processing}, pages 93--104, Brussels, Belgium, 2018.
  Association for Computational Linguistics.
\newblock \doi{10.18653/v1/D18-1009}.
\newblock URL \url{https://aclanthology.org/D18-1009}.

\bibitem[Zhang et~al.(2021)Zhang, Warstadt, Li, and Bowman]{zhang2021when}
Yian Zhang, Alex Warstadt, Xiaocheng Li, and Samuel~R. Bowman.
\newblock When {Do} {You} {Need} {Billions} of {Words} of {Pretraining} {Data}?
\newblock In \emph{Proceedings of the 59th {Annual} {Meeting} of the
  {Association} for {Computational} {Linguistics} and the 11th {International}
  {Joint} {Conference} on {Natural} {Language} {Processing} ({Volume} 1: {Long}
  {Papers})}, pages 1112--1125, Online, August 2021. Association for
  Computational Linguistics.
\newblock \doi{10.18653/v1/2021.acl-long.90}.
\newblock URL \url{https://aclanthology.org/2021.acl-long.90}.

\end{thebibliography}

\end{document}